\documentclass[10pt,twocolumn]{article}

\usepackage{iccv}
\usepackage{times}
\usepackage{epsfig}
\usepackage{graphicx}
\usepackage{amsmath}
\usepackage{amssymb}
\usepackage[english]{babel}
\usepackage[utf8]{inputenc}
\usepackage{xcolor}
\usepackage{multicol}
\usepackage[ruled,vlined]{algorithm2e}
\usepackage{caption}
\usepackage{tabularx}
\usepackage{bbm}
\usepackage[toc,page]{appendix}

\usepackage[pagebackref=true,breaklinks=true,colorlinks,bookmarks=false]{hyperref}

\iccvfinalcopy 

\def\iccvPaperID{9331} 

\usepackage[accsupp]{axessibility} 
\ificcvfinal\fi

\begin{document}

\title{SIMstack: A Generative Shape and Instance Model for Unordered Object Stacks}

\author{Zoe Landgraf$^{1}$, Raluca Scona$^{1}$, Tristan Laidlow$^{1}$, Stephen James$^{1}$ \\
Stefan Leutenegger$^{2}$ and Andrew J. Davison$^{1}$
\thanks{\small{Research presented in this paper has been supported by Dyson Technology Ltd.}}
\thanks{\small{$^{1}$Zoe Landgraf, Raluca Scona, Tristan Laidlow, Stephen James and Andrew J. Davison are with the Dyson Robotics Laboratory, Department of Computing, Imperial College London, UK.}
{\tt\small zoe.landgraf15@imperial.ac.uk}}
\thanks{$^{2}$Stefan Leutenegger is with the Smart Robotics Lab, Department of Computing, Imperial College London, UK.}
}

\maketitle
\ificcvfinal\thispagestyle{empty}\fi

\begin{abstract}
   By estimating 3D shape and instances from a single view, we can capture information about an environment quickly, without the need for comprehensive scanning and multi-view fusion. Solving this task for composite scenes (such as object stacks) is challenging: occluded areas are not only ambiguous in shape but also in instance segmentation; multiple decompositions could be valid. We observe that physics constrains decomposition as well as shape in occluded regions and hypothesise that a latent space learned from scenes built under physics simulation can serve as a prior to better predict shape and instances in occluded regions. To this end we propose SIMstack, a depth-conditioned Variational Auto-Encoder (VAE), trained on a dataset of objects stacked under physics simulation. We formulate instance segmentation as a centre voting task which allows for class-agnostic detection and doesn't require setting the maximum number of objects in the scene. At test time, our model can generate 3D shape and instance segmentation from a single depth view, probabilistically sampling proposals for the occluded region from the learned latent space. 
   Our method has practical applications in providing robots some of the ability humans have to make rapid intuitive inferences of partially observed scenes. We demonstrate an application for precise (non-disruptive) object grasping of unknown objects from a single depth view.
   
        
\end{abstract}

\vspace{-5mm}
\section{Introduction}

While humans are intuitively able to interpret partially observed scenes using geometric reasoning and prior experience, estimating 3D shape from RGB or depth images is challenging in computer vision due to ambiguity --- many 3D shapes can explain a 2D observation.
\begin{figure}
    \centering
    \includegraphics[width=\linewidth]{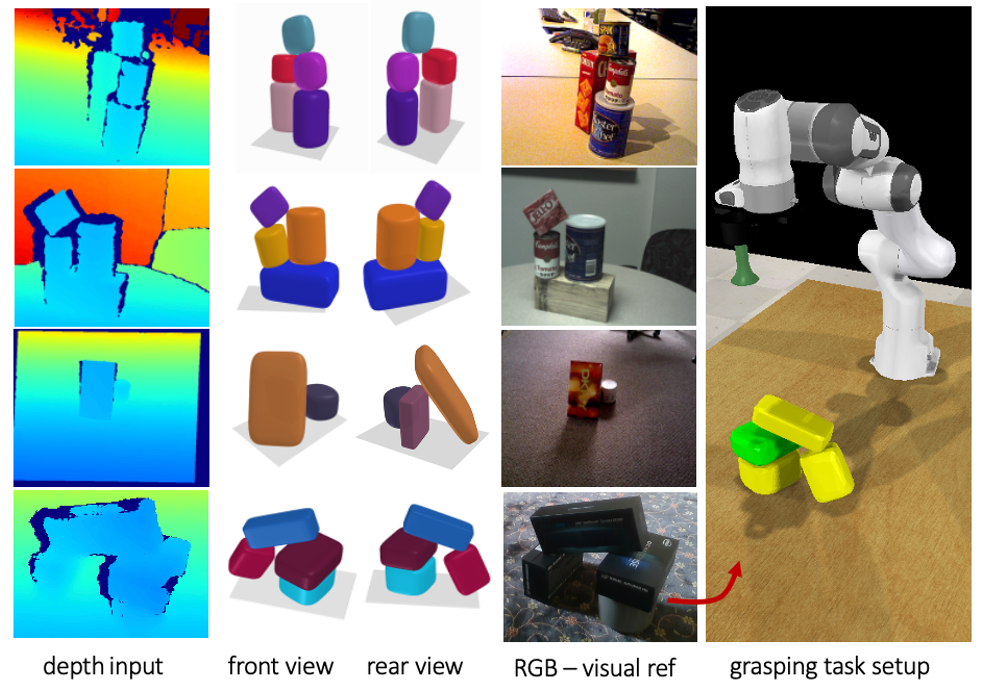}
    \caption{\small{\textbf{Top to bottom}: SIMstack outputs object shapes and instances from a single depth view on two YCB sequences and two real data examples, one with a fully occluded object supporting a leaning box ($3$rd row), for which our model predicts a plausible proposition. \textbf{Right}: Grasping demo setup (green: target object).}}
    \label{fig:Teaser_YCB_seq_examples}
\vspace{2mm} \hrule \vspace{-0.7cm} \end{figure}
The classical approach to generate 3D shapes from depth images involves taking images from all sides of an object and fusing the re-projected points into a common 3D representation such as a truncated signed distance function (TSDF) \cite{Newcombe:etal:ISMAR2011}. However, apart from the exhaustive nature of the task, it is often impossible to reach all required viewpoints to generate a watertight surface reconstruction \cite{Dai:etal:CVPR2018}.

This drawback has led researchers to explore learning based approaches to reconstruct 3D shapes, such as learning to complete partial reconstructions \cite{Dai:etal:CVPR2018, Dai_2019_CVPR, Stutz_2018_CVPR} and predicting scenes \cite{Song_2017_CVPR} or objects \cite{Yang_2017_ICCV,  8453803} from a single depth image. In parallel, researchers explore 3D shape prediction for scenes \cite{Nicastro_2019_ICCV,firman2016structured, Shin_2019_ICCV} and objects \cite{choy20163d,Xie_2019_ICCV,Yao_2020_CVPR}  from single or multi-view RGB data. While most of these approaches work with pointclouds, voxel occupancy grids or TSDF representations, some have explored alternative representations such as parametric surface elements \cite{Groueix_2018_CVPR}, 2D sketches \cite{marrnet}, graph neural networks \cite{Dai_2017_CVPR}, and the increasingly popular implicit neural surface representations \cite{Park_2019_CVPR, mildenhall2020nerf, Chabra2020DeepReconstruction}.

To our knowledge, no existing work has explored 3D shape prediction with instance segmentation for multiple objects from a single depth image.
Single view shape prediction allows to quickly estimate 3D occupancy, while instance segmentation is crucial for interactive tasks such as object manipulation. 
We aim to solve this task for a tabletop scene with a variable number of stacked household objects and propose a VAE whose latent space is
learned from stable (under physics simulation) scenes, with the aim to better reason about shape and instance decomposition in occluded regions. Conditioned on depth, the VAE learns to predict 3D shape and instances given a depth view, completing occluded regions using its latents, which can be thought of as an `intuitive physics' prior. Our refinement method further optimises the reconstruction for collision-free decomposition.
We aim to reconstruct scenes composed of unknown objects and train our model on randomly assembled piles of 3D parametric shapes (superquadrics). 
At test time, our conditional VAE (C-VAE) generates 3D shape and instance segmentation, as well as realistic reconstruction proposals for occluded regions in one forward pass. Our method can also integrate multiple views for improved reconstruction: the VAE can be conditioned on multiple views and our latent space can be further optimised against novel views using differentiable rendering. We show an application of our method for non-disruptive grasping using a robot arm. 

In summary, our contributions are:
\begin{itemize}
    \item A depth-conditioned VAE for scenes of stacked objects which can generate 3D shape, instance segmentation and probabilistic reconstruction proposals for occluded regions.
    \item A center voting scheme based on 3D Hough Voting allowing for class-agnostic 3D instance segmentation for scenes with an arbitrary number of objects.
    \item A shape refinement procedure to generate a compact scene representation of parametric shapes for downstream applications.
\end{itemize}
\section{Related Work}
\label{sec:Related_Work}
\vspace{-1mm}
\paragraph{Latent Representations for 3D Objects}
The shape of the occluded region of a 3D object is ambiguous and requires prior knowledge to estimate. 
Some methods use geometrical assumptions such as symmetry \cite{zhou2020learning}, but most tackle the problem using generative models and learn 3D shape priors using a latent representation \cite{Wu_2018_ECCV, Chen_2019_CVPR}. Zhang \etal \cite{NEURIPS2018_208e43f0} showed that such priors can generalize to unseen classes and Sucar \etal \cite{Sucar:etal:3DV2020} used class conditioned latent models to generate 3D object shapes by optimising the latent code against depth views. 
\vspace{-4mm}
\paragraph{Instance Segmentation of 3D Objects}
2D instance segmentation has progressed significantly with state-of-the-art methods such as Mask R-CNN \cite{He:etal:ICCV2017} and DETR \cite{10.1007/978-3-030-58452-8_13}, but segmenting objects in 3D remains challenging.
Hou \etal \cite{Hou_2019_CVPR} extend the idea of region proposals to 3D, combining backprojected 2D features with 3D partial scans. Another method for segmenting 3D pointclouds into objects is center voting \cite{Qi_2019_ICCV, Qi_2020_CVPR, Han_2020_CVPR}. We implement instance segmentation similarly, applying Hough Voting to 3D voxelgrids.
The instance segmentation approach of Xie \etal \cite{xie2020unseen} is closely related to ours --- their context is a cluttered table top and they use Hough Voting. However, their final output is a 2D instance segmentation whereas we aim at 3D output.
\vspace{-4mm}
\paragraph{Object Decomposition}
Previous work on decomposing compound objects into parts has focused on rigid objects such as furniture with significant structure and symmetry. Decomposing such objects into parts can help to leverage symmetry for shape completion  \cite{10.1145/2816795.2818094} or discover structure in unseen data \cite{Tulsiani_2017_CVPR}. Most approaches decompose into geometric primitives \cite{Deng_2020_CVPR,Tulsiani_2017_CVPR} using a library of shapes. Others use hierarchical representations and graph neural networks \cite{10.1145/3355089.3356527}. Paschalidou \etal \cite{Paschalidou_2019_CVPR} showed that using superquadrics over shape primitives improves reconstruction quality.
We model stacked objects as a compound object and leverage superquadrics to approximate a large variety of composing shapes. Unlike manufactured objects such as chairs, object stacks have no symmetry or class-specific structure. However, we show that their decomposition can be learned using a lower-dimensional representation.

\vspace{-1mm}
\section{Method}
\vspace{-1mm}
\label{sec:Method_Overview}
\paragraph{Task definition}
Given a single depth image of objects stacked on a tabletop, our task is to estimate the complete 3D shape of the group and segment the objects into instances. We learn a latent space that describes realistic pile configurations using a VAE, to improve segmentation and reconstruction in occluded regions. To allow the latents to focus on learning the occluded regions, we condition our VAE on features from one or multiple depth images.
Our goal is to train a generative model in the form of $p(x,\gamma,z) = p(z) p(x|z, \gamma) $ where $p(z)$ is a zero-mean multi-variate Gaussian prior describing realistic pile configurations. $p(x|z,\gamma)$ is the likelihood of pile configuration $x$ given latents $z$ and depth encoding $\gamma$ which we model with our depth conditioned, generative decoder.
We represent 3D geometry as a truncated signed distance function (TSDF) at a resolution of $64^\textbf{3}$, and instances using a center-voting vector field \ref{sec:instance_segmentation} at the same resolution.
\textbf{Pipeline} At test time our method takes as input one or more depth images and generates the complete TSDF of the object stack and a center-voting vector field, which, processed by our 3D Hough Voting algorithm, provides instance segmentation. We refine our output by fitting a superquadric to each predicted mesh (overview in Figure \ref{fig:arch}).


\subsection{A C-VAE for 3D Shape Instance Prediction}

\subsubsection{Network Architecture and training}
\label{sec:network_architecture}
\begin{figure*}[ht]
\centering{
\includegraphics[width=1.0\linewidth]{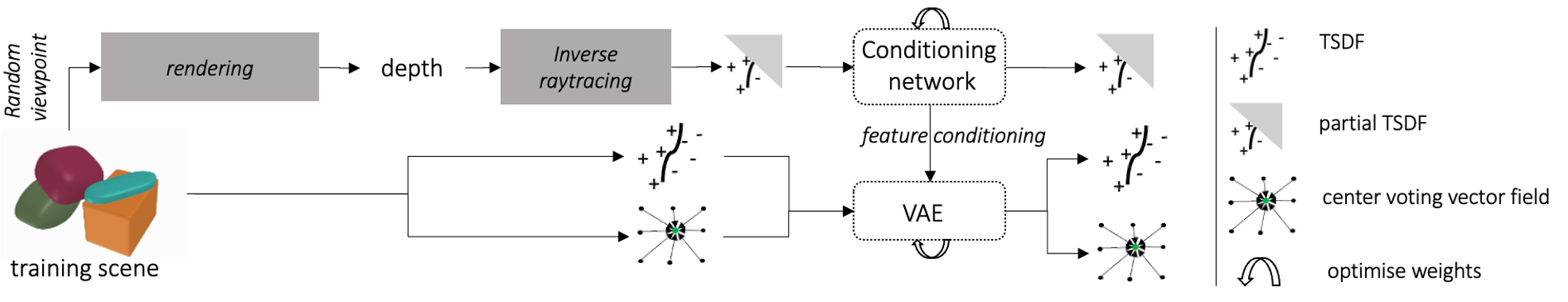}
}
\vspace{-0.8cm}
\caption{\small{Overview of our method (excluding shape fitting). Networks are trained jointly. Pre-processing steps (grey) are not optimised.}}
\label{fig:arch}
\vspace{2mm}
\hrule \vspace{-5mm}
\end{figure*}

Our network architecture consists of a 3D VAE which learns a latent space for realistic pile configurations and a conditioning network whose encoding layers feed their learned feature maps to the VAE, conditioning it on partial observations. Both encoder and decoder of the VAE split into task specific branches in the first and last layers, to allow for seperate encoding and decoding for instance and TSDF prediction. To learn descriptive depth features, the conditioning network is trained by auto-encoding a partial TSDF and is trained jointly with the VAE. We choose Residual blocks \cite{He:etal:CVPR2016} with Squeeze and Excitation (SE) \cite{Hu_2018_CVPR} over conventional convolutional layers in the joint encoder and decoder of the VAE to avoid gradient underflow and favour context integration. Since we compress both TSDF and instance segmentation into one latent code of size $96$, it learns a consistent joint  representation of the geometry and instance segmentation of a scene. Our architecture can be seen in Figure \ref{fig:arch}. \textbf{Training} To be effective, reconstruction has to be possible from any viewpoint. We therefore train our model on random viewpoints uniformly sampled during training in the range ($-1$m, $1$m) for lateral positions and ($10$cm, $1$m) for camera height. We generate a partial TSDF for every novel viewpoint as described in Section \ref{sec:conditioning_on_a_depth_image}). 
\vspace{-0.3cm}
\subsubsection{Conditioning on a Depth Image}
\label{sec:conditioning_on_a_depth_image}
Conditioning a 3D decoder on 2D depth image features would require the network to implicitly learn a reprojection task. To let it focus on the main task, we convert the input depth image(s) into a partial TSDF by reprojecting depth into the scene. We use inverse ray tracing for this and sample along each ray to fill the voxels in the camera view frustrum with the closest distance to a surface. Parallelized, our algorithm generates a partial TSDF in under $0.8$ seconds and can be used in online training.

\subsubsection{Loss Definition}
When predicting a TSDF, areas close to the surface with lower values will incur a smaller penalty than more distant positions. Other work \cite{NEURIPS2019_39059724} has proposed masking the loss by applying a different weight near the surface. We observe that this approach adds a discontinuity to the loss as well as an additional hyperparameter $\delta$ for the mask area. Instead we propose a loss which adds the inverse of the TSDF itself as a weighting factor:
\begin{eqnarray}
L_{\text{TSDF}} = {\sum_{x=0}^{X}\sum_{y=0}^{Y}\sum_{z=0}^{Z}{\frac{1}{|v_{x,y,z}| + \epsilon}  {|| v_{x,y,z} - \hat{v}_{x,y,z} ||_2}}},
\label{eq:distance_based_decay}
\end{eqnarray}
\noindent
where $v_{x,y,z}$ and $\hat{v}_{x,y,z}$ are the ground truth and predicted TSDF values at voxelgrid index $(x,y,z)$, $\epsilon$ prevents division by $0$ (set to $1e^{-9}$) and $|| . ||_2$ is the $L_{2}$-norm. We use the same loss for the partial TSDF prediction and refer to it as $L_{p\_{TSDF}}$.
For the instance vector voting task, we use the  $L_{2}$-norm between predicted and ground truth center vectors $\mathbf{\hat{\bar{c}}}$ and $\mathbf{\bar{c}}$:
\begin{eqnarray}
L_{\text{center\_votes}} &=& {\sum_{x=0}^{X}\sum_{y=0}^{Y}\sum_{z=0}^{Z}{|| \mathbf{\bar{c}}_{x,y,z} - \mathbf{\hat{\bar{c}}}_{x,y,z} ||_2}}.
\label{eq:instance_loss}
\end{eqnarray}
We found that this led to better convergence for the instance segmentation task than using cosine similarity. To approximate our prior, we empirically found that training using the Maximum-Mean Discrepancy $D_{MMD}$  \cite{zhao2017infovae} led to better convergence than using the standard KL-divergence \cite{Kingma:Welling:ICLR2014}. Our final loss is composed of $4$ components:
\begin{equation}
\begin{aligned}
    L_{total} = {} \alpha L_{TSDF} +  \beta L_{p\_{TSDF}} + \\ \gamma  L_{center\_votes} + \delta D_{MMD}.
\end{aligned}
\end{equation}
We set $\alpha=\beta=\gamma = 1$ and $\delta$ to $1e^5$.
\vspace{-0.2cm}
\subsubsection{A dataset of SuperQuadric Piles}
\label{sec:dataset}
\begin{figure}[ht]
\centering
\begin{minipage}[c]{0.99\linewidth}
\centering
  \includegraphics[width=0.999\linewidth]{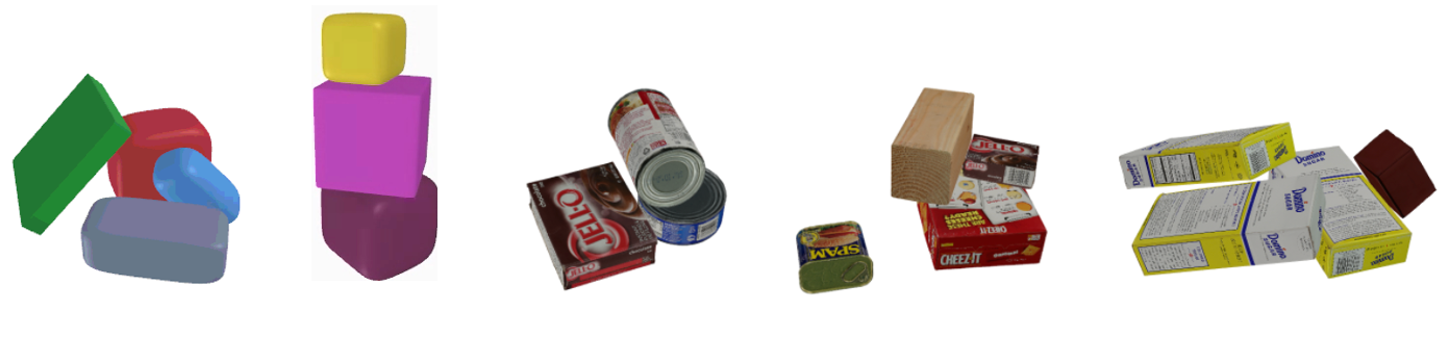}
\end{minipage}
  \caption{\small{\textbf{Left}: 2 object stacks from our Superquadric (SQ) training set. \textbf{Right}: 3 stacks from our YCB-object test set. All stacks are generated by random placement  under physics simulation. }}
\label{fig:SQ_dataset}
\vspace{2mm} \hrule \vspace{-0.7cm} \end{figure}

Instead of training on specific shapes from datasets such as ShapeNet \cite{Shapenet:ARXIV2015} or ModelNet \cite{Wu:etal:CVPR2015}, we use generic shapes to favour generalisation. Superquadrics offer a general shape description, extending quadrics to multiple exponents, and are defined as the solution to the implicit equation: 
\begin{equation}
\label{eq:sq}
    f(x, y, z) = \left| \frac{x}{a_1} \right|^{r} + \left| \frac{y}{a_2} \right|^{s} + \left| \frac{z}{a_3} \right|^{t} = 1.
\end{equation}
For exponents $r$, $s$, $t >= 1$ the shapes are convex and varying the scale parameters $a_1$, $a_2$, $a_3$  generates very flat, small, or bulky shapes, ideal for approximating everyday household objects. We leverage this continuous parametrisation to sample a large variety of shapes ($3500$ individual shapes with exponents ranging between $2$ to $100$ and scales between $5$ cm and $30$ cm) and generate $10,000$ realistic object piles using PyBullet \cite{coumans2018pybullet}. For each scene we randomly place between $3$ and $4$ superquadrics (see Figure \ref{fig:SQ_dataset}) .

\subsection{Postprocessing}
\subsubsection{Class-agnostic Instance Segmentation}
\label{sec:instance_segmentation}
\begin{figure}[ht]
\centering{
\begin{minipage}[c]{0.99\linewidth}
\centering
  \includegraphics[width=0.9\linewidth]{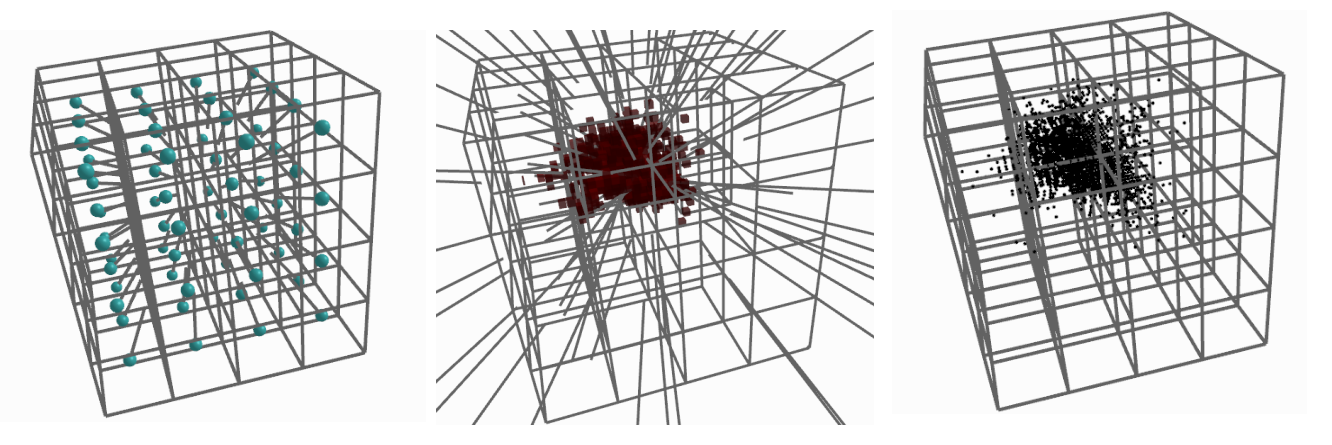}
\end{minipage}
}
  \caption{\small{Our Hough Voting method illustrated for a single object center. \textbf{Left}: The network output --- unit vectors pointing towards the center of the object. \textbf{Center}: Raycasting and extracting those voxels with a high number of traversal at a 10-fold voxelgrid resolution. \textbf{Right}: Voxels with a high ray-traversal rate are likely close to the object center and selected as 'votes'. To obtain final centers, selected votes are further processed using Mean shift clustering.}}
\label{fig:HoughVoting}
\vspace{1.5mm} \hrule \vspace{-0.7cm} \end{figure}

To make our instance segmentation class-agnostic and independent of the number of objects present, we implement instance segmentation using 3D Hough voting. The instance branch of our network
predicts a vector field in which each voxel `votes' for the object it belongs to by predicting a 3D unit vector $\mathbf{\hat{c}}$ from its own centroid to the object's centroid. (see Figure \ref{fig:HoughVoting}). At a 10-fold higher resolution than our voxel grid $(640 \times 640)$ , we count the number of rays that traverse each voxel by marching each ray through the grid. We use the object bounds obtained from the TSDF prediction to limit raycasting to the inside of objects. A higher resolution allows for more detailed prediction of centers which don't coincide with voxel centers at the original resolution. Those voxels with a number of traversals larger than $\mu$ are selected as object center votes and passed to MeanShift to obtain the final center locations. To allocate voxels to their corresponding centers, we compute:
\begin{equation}
    \min_{c_{1}, \dots c_{N}} {|\arccos(\mathbf{\hat{c}} \cdot \mathbf{\hat{c_{n}})| + \gamma * \|\mathbf{\hat{c_{n}}} \|}}
\end{equation}
whereby $\mathbf{\hat{c_{n}}}$ is the normalised distance from the voxels' center to cluster center $n$. Setting hyperparameters $\mu = 10$ and $\gamma = 0.1$ gave the best performance.  While we observe limited sensitivity to $\gamma$, very large ($\geq25)$ and small ($\leq4$) values of $\mu$ increase under- and over-segmentation respectively.  Paralellized on the GPU our method takes on average $19 ms$ for a scene resolution of $64^{3}$.

\vspace{-2mm}
\subsubsection{Shape Refinement using Superquadric Fitting}
\label{sec:SQ_fitting}
Our system outputs a TSDF and voxelised instance segmentation of the compound object; if needed, additional post-processing can be applied (e.g. primitive shape fitting, CAD-model fitting). To generate a compact representation, we propose following refinement procedure, which extracts a set of superquadrics and their poses from the raw output:

First, we run marching cubes on the TSDF and sample from the resulting mesh to get a point cloud.
We segment this point cloud according to the voxelised instance predictions.
We then independently fit a superquadric to the $N_i$ points belonging to the $i$th segment by minimising the ``mean distortion'' \cite{Chevlaier:etal:2003}, $L_{\text{dist}, i}$, given by:
\begin{equation}
    L_{\text{dist}, i} = {\frac{1}{N_i} \sum_{k=1}^{N_i} \sqrt{a_{i1} \: a_{i2} \: a_{i3}} \cdot d_i(x_{i, k}, y_{i, k}, z_{i, k})^2},
\end{equation}
\noindent
where
\begin{equation}
     d_i(x_{i, k}, y_{i, k}, z_{i, k}) = \Vert \overrightarrow{O_iP_k} \Vert \frac{f_i(x_{i, k}, y_{i, k}, z_{i, k}) - 1}{f_i(x_{i, k}, y_{i, k}, z_{i, k})},
\end{equation}
\noindent
$a_{i1}$, $a_{i2}$, $a_{i3}$ are the scale parameters and $f_i(x_{i, k}, y_{i, k}, z_{i, k})$ is the implicit equation from Eq.~\ref{eq:sq}, $d_i(x_{i, k}, y_{i, k}, z_{i, k})$ is the approximate Euclidean distance to the surface \cite{Bardinet:etal:1995}, $\Vert \overrightarrow{O_iP_k} \Vert$ is the distance from the point to the superquadric center.

We then optimise for the parameters of all $Q$ superquadrics together, with additional cost terms to penalise collisions between superquadrics and intersection with the floor plane:
\begin{equation}
    L = \sum_i^Q \left[ \lambda_\text{dist} L_{\text{dist}, i} + \lambda_\text{coll} \sum_{\genfrac{}{}{0pt}{}{j=1}{i \neq j}}^Q L_{\text{coll}, ij} + \lambda_\text{floor} L_{\text{floor}, i} \right].
\end{equation}
\noindent
The cost of sample points from superquadric $j$ colliding with superquadric $i$, $L_{\text{coll}, ij}$, is given by:
\begin{equation}
    {L_{\text{coll}, ij} = \frac{1}{N_j} \sum_{k=1}^{N_j} \mathbbm{1}_{d_i(x_{j, k}, y_{j, k}, z_{j, k}) < 0} \cdot d_i(x_{j, k}, y_{j, k}, z_{j, k})^2},
\end{equation}
\noindent
where $\mathbbm{1}$ is the indicator function, and the cost of sample points from superquadric $i$ colliding with the floor plane, $L_{\text{floor}, i}$, is given by:
\begin{equation}
    {L_{\text{floor}, i} = \mathbbm{1}_{z_{i, k} < 0} \cdot z_{i, k}^2.}
\end{equation}
Experimentally, we found $\lambda_\text{dist} = 100$, $\lambda_\text{coll} = 10$, and $\lambda_\text{floor} = 1$ to work well.
For both stages, we minimised the cost function using the dogleg algorithm with rectangular trust regions, bounding the superquadric parameters to enforce convex shapes.

\subsection{Multi-View Estimation}
\label{sec:optimising_the_latent_space}
Our method supports multi-view reconstruction in two ways: the generative decoder can be conditioned on a partial TSDF generated from fusing multiple depth images (\textit{multi-view conditioning}); or the latent code can be optimised from one or multiple depth views using differentiable depth rendering (\textit{multi-view optimisation}). To implement multi-view conditioning, we backproject multiple depth views into the scene using inverse raytracing as in Section \ref{sec:conditioning_on_a_depth_image}. To demonstrate how our latent code can be optimised against multiple depth views, we implement a differentiable depth renderer for raytracing TSDFs and an optimization method described in the Appendix B.

\section{Experiments}
\begin{figure}[ht]
\centering
\begin{minipage}[c]{0.99\linewidth}
\includegraphics[width=\linewidth]{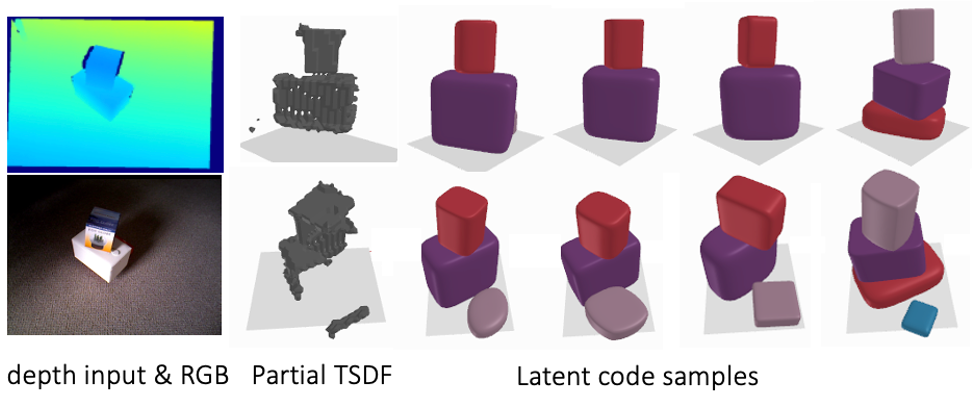}
\end{minipage}
\caption{\small{Sampling from the latent space for a real data example. We see multiple pile samples which all look similar from the point of view of the input depth observation (\textbf{top}) but significantly different though all plausible from the back (\textbf{bottom}).}}
\label{fig:Occlusion_proposals_example} \vspace{2mm} \hrule \vspace{-0.4cm} 
\end{figure}

\begin{figure}[ht]
\centering
\begin{minipage}[c]{1.0\linewidth}
\centering
\includegraphics[width=1.0\linewidth]{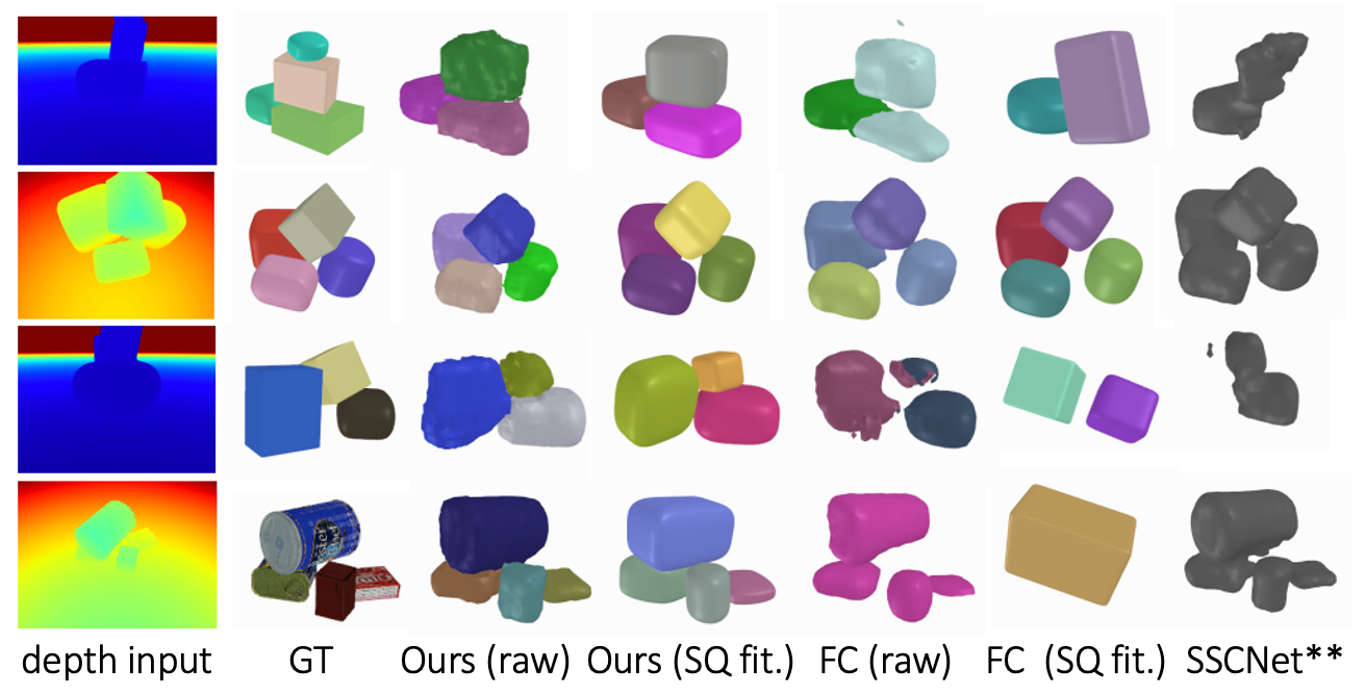}
\end{minipage}
    \vspace{-0.2cm} 
  \caption{\small{Comparing our method against our baselines on examples from our SQ test dataset and our YCB test dataset. More qualitative examples are shown in Appendix A.}}
\label{fig:Qual_comparison_to_baselines}
\vspace{2mm} \hrule \vspace{-0.2cm} 
\end{figure}

We evaluate our method quantitatively on our own test dataset ($1419$ scenes) and a test dataset of YCB  \cite{Calli:etal:ICAR2015} objects ($256$ scenes) to demonstrate generalisation. For the latter we select YCB objects with IDs: \textit{2, 3, 4, 5, 7, 8, 9, 10, 36, 61} 
which can be approximated by a single superquadric and generate object stacks using the same procedure we used for our own synthetic scenes (Section \ref{sec:dataset}). We show dataset examples in Figure \ref{fig:SQ_dataset}.
We demonstrate qualitative result on our synthetic datasets in Figure \ref{fig:Qual_comparison_to_baselines} as well as on real data examples and sequences of the YCB-Video dataset (Figures \ref{fig:Teaser_YCB_seq_examples},
\ref{fig:real_data_ex} and  \ref{fig:Occlusion_proposals_example}) and evaluate our method for multi-view reconstruction. We show how our method generalises to scenes with more objects and demonstrate an application of our method for non-disruptive grasping on unknown objects. 

\subsection{Single-view Reconstruction}
\label{sec:single_view_reconstruction}
We are not aware of directly comparable work on 3D instance segmentation for object groups and therefore select the following two baselines to compare our method against.
\textbf{SSCNet Baseline}
The closest existing method is SSCNet \cite{Song_2017_CVPR}, which predicts a 3D occupancy grid with semantic labels for multi-object rooms from a single depth image. As this method does not predict instances, we use it as a baseline only for geometric reconstruction  and adapt it as follows: (1) we change the last layer to predict a TSDF; (2) we remove downsampling (required only in large-scale scenes) from layers by adding padding as well as one upsampling layer.
We refer to this modified version as $SSCNet^{**}$.
\textbf{Fully Convolutional (FC) Baseline}
Secondly, we use an ablated version of our model: a fully convolutional network predicting a TSDF and instance vector field from a single depth image. This serves both as a baseline for instance segmentation and an ablation study, showing the advantage of a learned prior over direct prediction. Our ablated model uses the same SE and Residual units as our main model
but no feature space compression, as we found it reduced convergence. A detailed description of baseline architectures can be found in Appendix C.

\begin{figure}[ht]
\centering{
\begin{minipage}[c]{0.35\linewidth}
\begin{minipage}[c]{0.65\linewidth}
  \includegraphics[width=\linewidth]{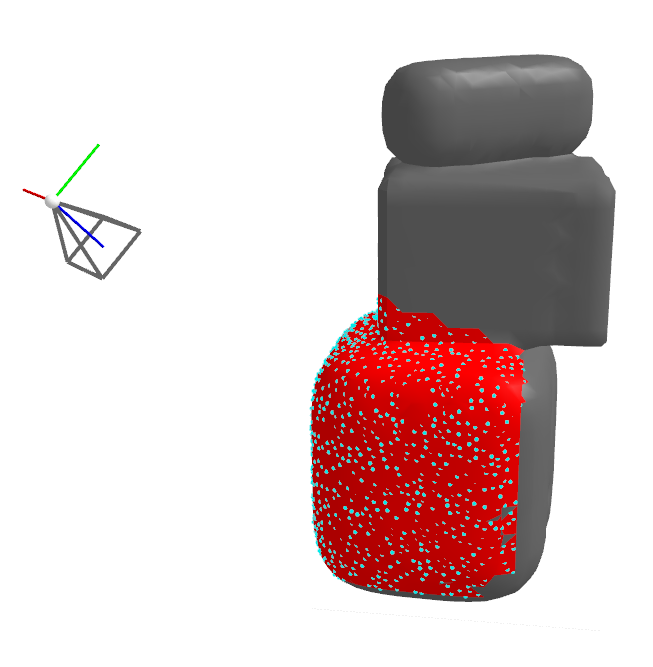}
\end{minipage}
\end{minipage}
}
\begin{minipage}[c]{0.45\linewidth}
\centering{
\begin{minipage}[c]{0.3\linewidth}
\includegraphics[width=\linewidth]{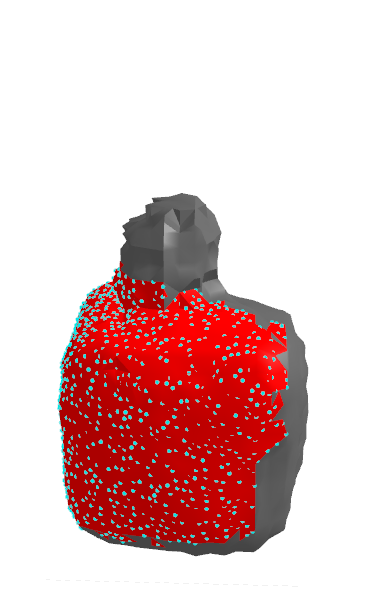}
\end{minipage}
\begin{minipage}[c]{0.3\linewidth}
\includegraphics[width=\linewidth]{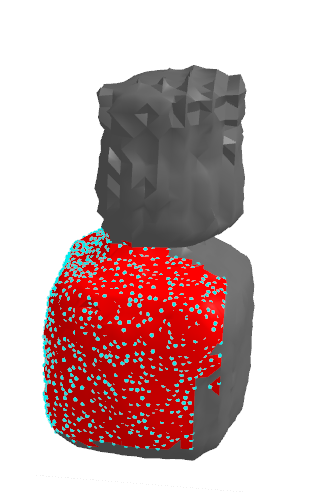}
\end{minipage}
\begin{minipage}[c]{0.3\linewidth}
\includegraphics[width=\linewidth]{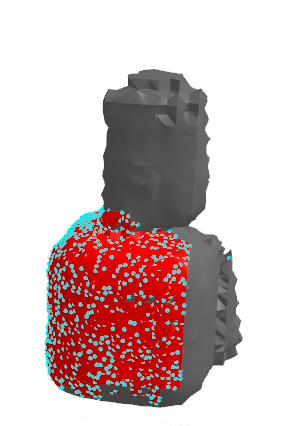}
\end{minipage}
}
\end{minipage}
  \caption{
  \small{\textbf{Left}: Ground truth with the visible part of the mesh highlighted in red.
  \textbf{Right}: 3 random latent code samples generated by our network.  Note how every sample has different reconstructions in the occluded areas, which are plausible but would yield high reconstruction error with respect to the ground truth TSDF.}}
 \vspace{-2mm}
\label{fig:evaluation_method_vis}
\vspace{2mm} \hrule \vspace{-0.5cm} \end{figure}
\vspace{-3mm}
\subsubsection{Reconstruction Accuracy}
\label{sec:reconstruction_acc} 
\begin{figure}[ht]
\begin{minipage}[c]{0.46\linewidth}
  \includegraphics[width=\linewidth]{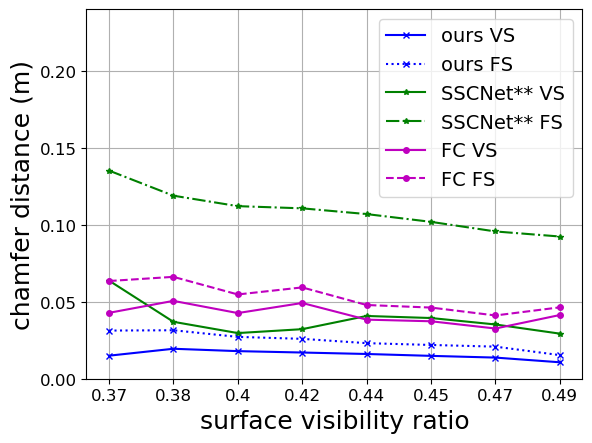}
\end{minipage}
\begin{minipage}[c]{0.46\linewidth}
  \includegraphics[width=\linewidth]{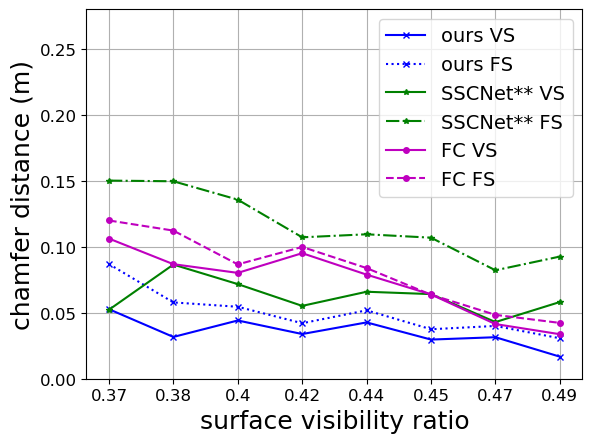}
\end{minipage}
  \caption{\small{Chamfer Distance for 1000 samples on our test datasets of SuperQuadric shapes (\textbf{left}) and object stacks composed of YCB objects (\textbf{right}). We plot results for different surface mesh visibility ratios. \textbf{VS}: Visible surface. \textbf{FS}: Full surface}}
\label{fig:CD_test_and_YCB}
\vspace{2mm} \hrule \vspace{-0.2cm} \end{figure}

Comparing the full reconstruction to the ground truth scene unfairly penalises the network in occluded regions for reconstructions which are plausible, but differ from the ground truth (see Figure \ref{fig:evaluation_method_vis}). We therefore evaluate our method on the visible surface area and the full reconstruction separately.
To show how performance differs depending on how much of the object stack is visible from a given viewpoint, we report results by surface visibility. 
Note that given a single view, the maximum surface visibility of a pile is around $50\%$, since the back will always be occluded. Since our method is viewpoint agnostic, we evaluate every test scene from $3$ random viewpoints uniformly sampled from the same ranges used during training (Section \ref{sec:network_architecture}). We ensure the distance between camera and scene centers is at least $50cm$. For every viewpoint, we generate a 3D scene and compute the average reconstruction accuracy for $3$ latent code samples to account for variability. 
\textbf{Visible surface evaluation} To evaluate surface reconstruction accuracy, we extract a mesh from the generated and ground truth TSDFs using marching cubes. We obtain the visible surface by extracting all visible faces using ray-triangle intersection with the generated mesh. 
We produce $1000$ uniform samples from the extracted surface and evaluate the bidirectional Chamfer Distance between ground truth and predicted point sets, $P_{m}$ and $P_{k}$:
\begin{equation}
    \textit{CD} = {\frac{1}{N_{m}}\sum_{x_{i}=0}^{N_{m}}{\min_{x_{j} \epsilon P_{k}}(\mathbf{x_{i}} - \mathbf{x_{j}})} + \frac{1}{N_{k}}\sum_{x_{i}=0}^{N_{k}}{\min_{x_{i} \epsilon P_{m}}(\mathbf{x_{j}} - \mathbf{x_{i}})}}
    ~,
\label{eq:CD}
\end{equation}
where $\mathbf{x_{i}}$ and $\mathbf{x_{j}}$ are the 3D coordinates of sampled points from $P_{m}$ and $P_{k}$ respectively. 
\textbf{Full reconstruction evaluation}
We evaluate the full reconstruction in terms of surface reconstruction accuracy and predicted voxel occupancy. For the former we compute the Chamfer Distance between $1000$ samples from the full predicted and ground truth mesh surfaces. For the latter we compute the Binary Cross Entropy (BCE):
\begin{equation}
    -\frac{1}{N} \sum_{\Omega} {y_{i} \log({\frac{1}{S}\sum_{j=0}^{j=S}{\hat{y_{j}}}}) + (1-y_{i}) \log({1 - \frac{1}{S}\sum_{j=0}^{j=S}{\hat{y_{j}}}})}
    ~,
    \label{eq:binary_cross_entropy}
\end{equation}
where $N$ is the voxelgrid resolution $64^3$, $S$ is the number of latent code samples taken per viewpoint, which we set to $3$ and $\hat{y}$ and y are the predicted and ground truth occupancy value respectively, computed as follows:
\begin{eqnarray}
  Occupancy_{x,y,z} = \begin{cases}
       {0} \quad\text{if } TSDF_{x,y,z} > 0 \\
       {1} \quad\text{if } TSDF_{x,y,z} <= 0
     \end{cases}
     \label{eq:TSDF_to_OCC}
\label{eq:occupancy_from_TSDF}
\end{eqnarray}
We also report the Intersection over Union, whereby we use a threshold of $0.5$ to distinguish between (label) occupied and unoccupied space. We evaluate our baselines using the same procedure and display our results in Figure \ref{fig:CD_test_and_YCB} and Table \ref{tab:results}. Note that we set $S$ to 1 in (\ref{eq:binary_cross_entropy}) for our baselines and just use the prediction $\hat{y}$. Our method outperforms both baselines for full reconstruction (Chamfer Distance evaluated for the full mesh surface, and expected occupancy) and visible surface reconstruction, showing its overall advantage for reconstructing and decomposing a 3D scene.
Note that the gap between visible and full surface reconstruction is largest for $SSCNet^{**}$, but more comparable between the FC baseline and our method. As expected for all methods, with lower visibility full surface reconstruction accuracy drops faster than visible surface accuracy. The drop is however most pronounced for $SSCNet^{**}$. These observations suggest that while using a prior improves overall performance, a large improvement in predicting occluded space is achieved by jointly predicting shape and instances.
\vspace{-5mm}
\subsubsection{Scene decomposition (instance segmentation)}
Similarly to the reconstruction accuracy, it is impossible to make an exact quantitative evaluation of our method's instance segmentation since decomposition estimates in occluded regions can be plausible even when very different from the ground truth. In Figure \ref{fig:Occlusion_proposals_example}, for example, our method hypothesises plausible hidden objects, although the ground truth does not contain one. We therefore decided to evaluate instance segmentation as a valid scene decomposition which generates stable piles in physics simulation.
\textbf{Stability evaluation under physics simulation}
We evaluate stability by loading our generated 3D meshes into the PyBullet physics engine and simulating 10000 steps with a gravity setting of of $10 \frac{m}{s^2} $ along $-z$, a friction coefficient of $1$ and assuming uniform density. We compare our method against our FC baseline (which provides instance segmentation) in terms of object center displacement after simulation and report results in Table \ref{tab:results}. Our method outperforms the fully convolutional baseline by over $50cm$ on both our test datasets indicating that the collections generated by our method are generally more stable and hence plausible. Note that in the case of rolling objects as well as intersecting objects (causing objects to be pushed apart due to contact force), average object displacement can become large.
\begin{figure}[ht!]
\begin{minipage}[c]{1.0\linewidth}
\centering{
\includegraphics[width=\linewidth]{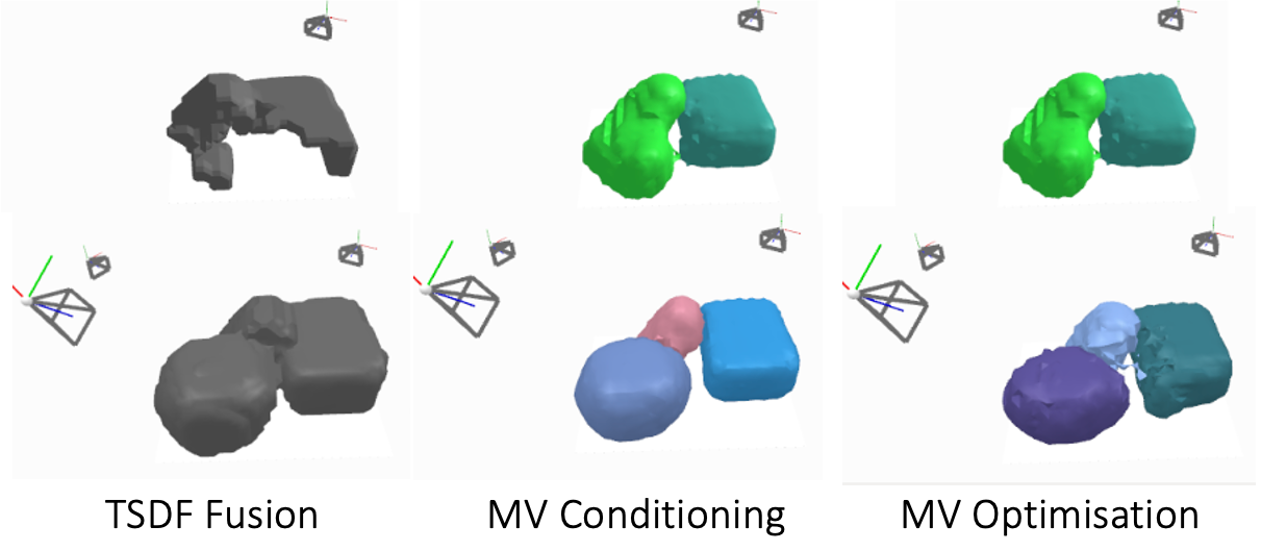}
}
\end{minipage}
\vspace{-0.5cm} 
 \caption{\small{Example of multi-view reconstruction comparing multi-view conditioning (MV cond), multi-view optimisation (MV opt) to TSDF-Fusion. We show raw results (no SQ fitting)}}

\label{fig:Multi_view_Comparison}
\vspace{2mm} \hrule \vspace{-0.2cm} 
\end{figure}

\vspace{-0.7cm} 
\subsubsection{Evaluating on Real Data}
\begin{figure}[ht]
\centering{
\begin{minipage}[c]{0.95\linewidth}
\includegraphics[width=\linewidth]{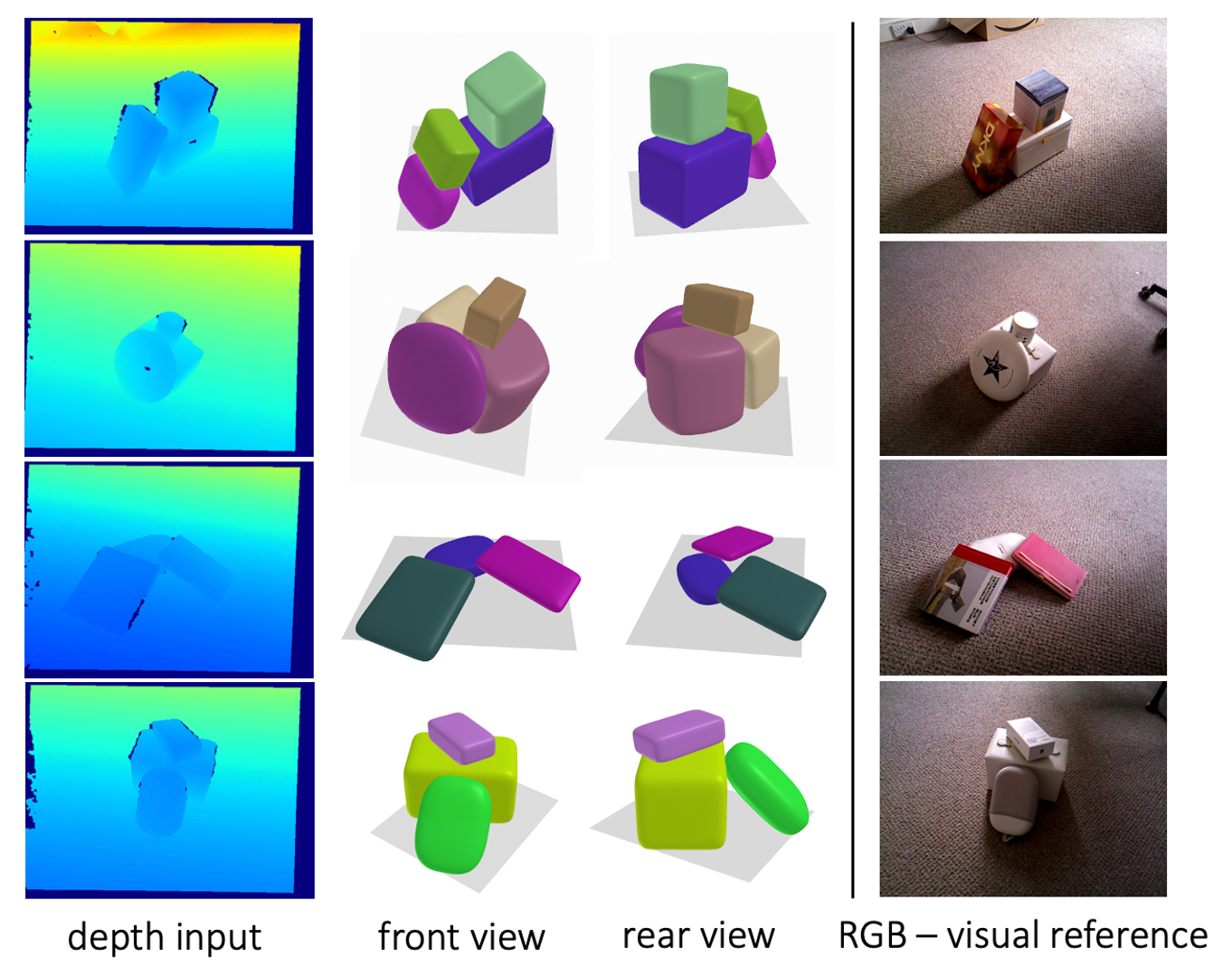}
\end{minipage}
}
  \caption{\small{Qualitative results on real data examples. Our method generates 3D shape and instances from single depth views.}}
\label{fig:real_data_ex}
\vspace{0.5mm} \hrule \vspace{-0.5cm} 
\end{figure}

We qualitatively evaluate the proposed method on real data. We collect RGB-D images of structures (leveraging ORB-SLAM \cite{Mur-Artal:etal:TRO2017} poses) and post-process them by (1) segmenting the floor plane using RANSAC \cite{Rusu:etal:ICRA2011} and (2) rectifying the roll and pitch of the camera pose by aligning the floor plane normal with the vertical axis. 
Our results (Figure \ref{fig:real_data_ex}, \ref{fig:Teaser_YCB_seq_examples} and \ref{fig:Occlusion_proposals_example}) show that our method generates correct 3D reconstruction, realistic occlusion proposals and instance segmentation for a number of real world examples. Our example in Figure \ref{fig:Occlusion_proposals_example} shows that our method realistically fills in the occluded area below the box by extending its shape to the floor. When sampling from the latents, it is able to hypothesise different occluded objects. In particular, our example in Figure \ref{fig:Teaser_YCB_seq_examples} ($3$rd row) shows that our network is capable to hypothetise a supporting object which is completely occluded but required to support a leaning box. We also evaluate our method on YCB-Video sequences (Figure \ref{fig:Teaser_YCB_seq_examples}).

\begin{table}
\begin{center}
\resizebox{0.48\textwidth}{!}{%
\begin{tabular}{|c||c|c|c|c|c|c|}
 \hline
\multicolumn{1}{|c|}{} & \multicolumn{2}{|c|}{\textbf{ours}} & \multicolumn{2}{|c|}{\textbf{FC}}  & \multicolumn{2}{|c|}{\textbf{SSCNet**}} \\
 \hline
 \multicolumn{7}{|c|}{\small{Avg. Chamfer Distance (m)}} \\
 \hline
  & ($\mu$) & ($\sigma$) & ($\mu$) & ($\sigma$)  & ($\mu$) & ($\sigma$)\\ 
 \hline
 \small{visible surface (SQ dataset)} & 0.016 & 0.0002 & 0.044&  0.002& 0.036 & 0.002\\ 
 \hline
 \small{full surface (SQ dataset)}  & 0.028 & 0.0003 & 0.062 &0.107 &  0.003 &  0.002\\ 
 \hline
 \small{visible surface (YCB dataset)} & 0.036  & 0.002 & 0.076 & 0.006 & 0.065 & 0.008 \\ 
 \hline
 \small{full surface (YCB dataset)}  & 0.062 & 0.007 & 0.098 & 0.011 & 0.116 & 0.009 \\ 
 \hline
\multicolumn{7}{|c|}{\small{Expected occupancy }} \\
 \hline
\multicolumn{1}{|c||}{} & \multicolumn{6}{|c|}{\small{Binary Cross Entropy}} \\
 \hline
 SQ dataset & 0.089 & 0.003 & 0.175 & 0.005 & 0.247 & 0.008\\ 
 \hline
  YCB dataset & 0.112 & 0.006 & 0.236 & 0.029 & 0.242 & 0.015\\ 
 \hline
 \multicolumn{1}{|c||}{} & \multicolumn{6}{|c|}{\small{Intersection over Union}} \\
 \hline
 SQ dataset & 0.829 & 0.004 & 0.635 & 0.005 & 0.424 & 0.004\\ 
 \hline
  YCB dataset & 0.784 & 0.003 & 0.542 & 0.012  & 0.460 & 0.006 \\ 
 \hline
 \multicolumn{7}{|c|}{\small{Avg. object centre displacement (m)}} \\
 \hline
 SQ dataset & 0.593 & 1.28 & 1.112 & 5.441 & -- & --\\ 
 \hline
  YCB dataset & 1.216 & 4.734 & 1.756 & 6.494 & -- & -- \\ 
 \hline
\end{tabular}}
\end{center}
\vspace{-0.5cm} 
\caption{\small{Comparison in terms of Chamfer Distance, expected occupancy and stability under physics simulation. We report the average results for $3$ viewpoints per scene.}}
\vspace{2mm} \hrule \vspace{-0.2cm} 
\label{tab:results}
\vspace{-5mm}
\end{table}

\begin{figure*}[ht!]
\centering
\includegraphics[width=1.0\linewidth]{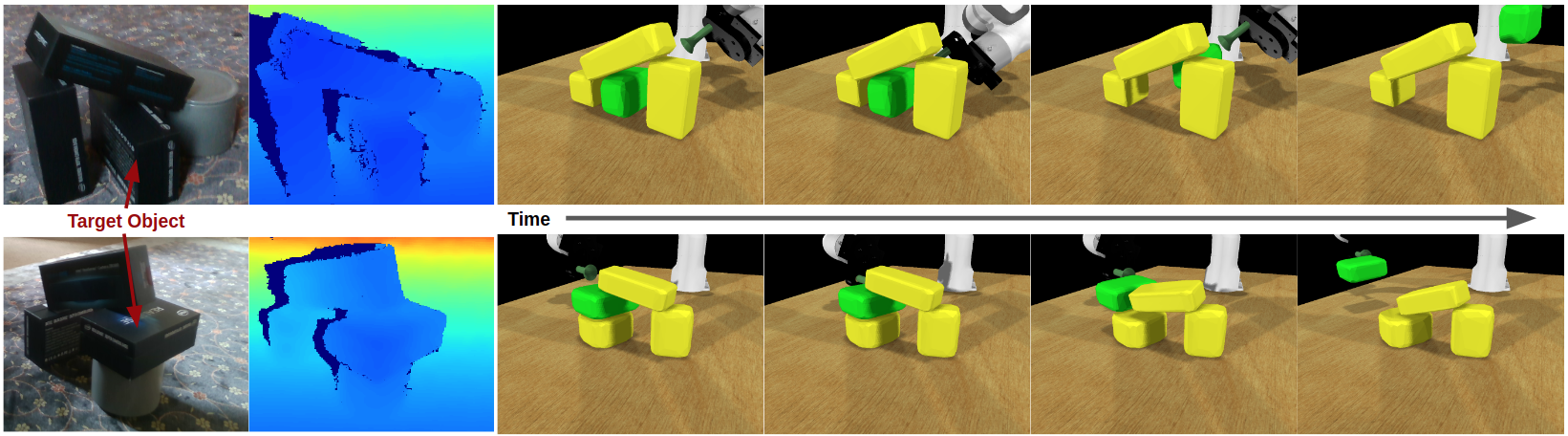}
\caption{\small{Two examples from our grasping system. Target objects are highlighted in green. The first example shows the robot sliding the target object sideways, leaving the other objects undisturbed. The second example grasps the object from the side and pulls it out such that the top resting object slides onto a lower supporting object, causing minimum disruption to the stack.}}
\label{fig:robot_demo}
\vspace{1mm} \hrule \vspace{-3mm}
\end{figure*}

\begin{figure}[ht]
\centering{
\begin{minipage}[c]{0.99\linewidth}
\includegraphics[width=\linewidth]{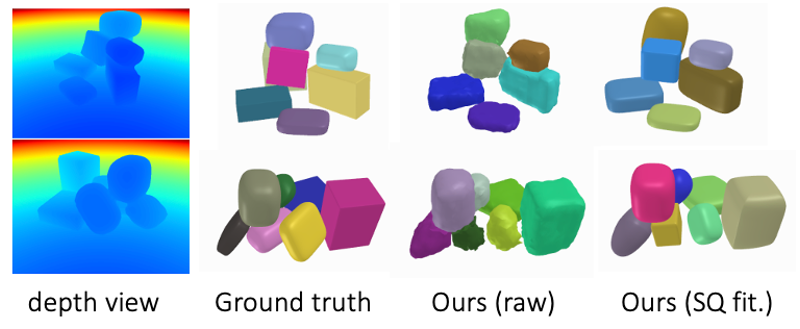}
\end{minipage}
}
  \caption{\small{Qualitative results on  scenes with $7$ objects, demonstrating the ability of our method to generalise to more objects.}}
\label{fig:more_objects}
\vspace{2mm} \hrule \vspace{-0.3cm} 
\end{figure}

\subsection{Multi-View Reconstruction}
\label{sec:evaluating_multi_view_reconstruction}
\begin{figure}[ht]
\centering{
\begin{minipage}[c]{0.45\linewidth}
  \includegraphics[width=\linewidth]{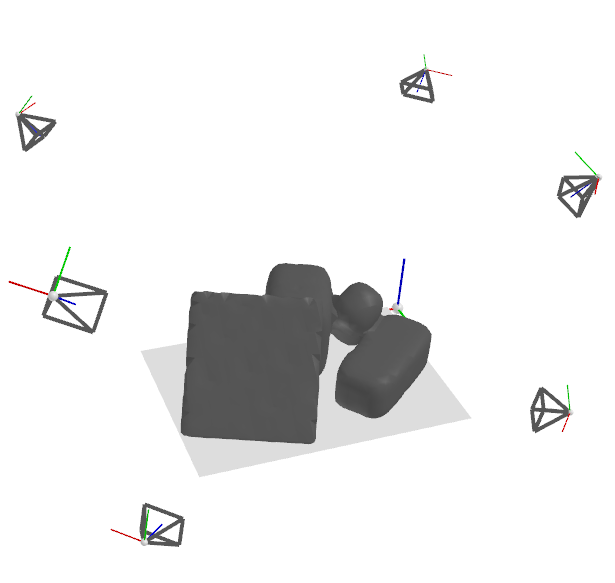}
\end{minipage}
}
\centering{
\begin{minipage}[c]{0.45\linewidth}
  \includegraphics[width=\linewidth]{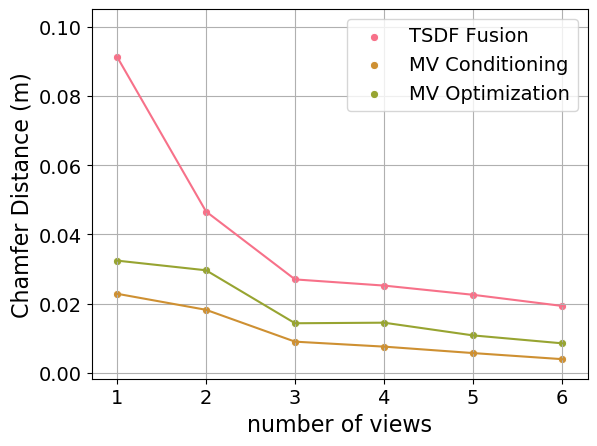}
\end{minipage}
}
 \caption{\small{Multi-view estimation. \textbf{Left}: The 6 viewpoints for which we evaluate. \textbf{Right}: Chamfer Distance (1000 points) for increasing views of TSDF Fusion, multi-view conditioning (MV Conditioning) and multi-view optimisation (MV Optimisation).}}
\label{fig:Multi_view_quantitative_comparison}
\vspace{2mm} \hrule \vspace{-0.7cm} \end{figure}

We evaluate the performance of our method using multi-view conditioning and multi-view optimisation. We use simple TSDF Fusion as a baseline which we implement using the method described in \ref{sec:conditioning_on_a_depth_image}.  For a fixed sequence of 6 viewpoints (see Figure \ref{fig:Multi_view_quantitative_comparison}) we compare the reconstruction quality for our SQ test dataset. For multi-view conditioning, we average reconstruction accuracy over 2 latent code samples. Note that the first view covers on average $30\%$ of the test scenes. 
For multi-view optimization, we condition on the first view and optimise the latent code against the additional views (see Appendix B for details). Our experiments show that multi-view conditioning generates the best reconstruction and that both our multi-view methods outperform TSDF Fusion, even at very high visibility. We attribute this to the fact that our generative method also reconstructs the bottom of objects, which will always be occluded and therefore cannot be reconstructed by TSDF Fusion. Our example in Figure \ref{fig:Multi_view_Comparison} shows how a scene reconstruction is updated with additional views. Please refer to Appendix D and the \href{https://zoelandgraf.github.io/SIMstack/}{supplementary video} for additional qualitative examples and runtime analysis. 
\vspace{-3mm}
\subsubsection{Additional qualitative experiments}
\textbf{More objects and non-convex objects}  Although only trained on 3-4 objects, SIMstack generalises well to scenes with up to $7$ objects (see Figure \ref{fig:more_objects}). We attribute this to the fact that our formulation of instance segmentation is independent of the number of objects in the scene and that our model learns from scenes with varying number of objects. Our method also shows some ability to generalise to non-convex objects (see Appendix D for examples).  \textbf{Latent code analysis} Our multi-view optimization experiments show that our latent code can be optimised against novel views, generating consistent shape and instance predictions (Figure \ref{fig:Multi_view_Comparison}). This suggests that our shape encoding is smooth as well as consistent between shape and instance representation. We provide further qualitative examples demonstrating this in Appendix D.


\subsection{Precise Robot Manipulation}

Robotic grasping is a well studied problem~\cite{bohg2014data} and an active area of research~\cite{pinto2016supersizing, mahler2017dex, james2019sim}. However, many state-of-the-art grasping systems perform indiscriminate grasping, with no regard of how the grasp might effect surrounding objects. 
However, multi-object reasoning is 
essential when grasping in cluttered scenes; grasping in an imprecise manner may topple a stack and cause damage to fragile objects. In this section, we show how SIMstack can be used to perform precise 6D grasping of a target object while minimising disruption to surrounding objects.

Our grasping demo consists of a real stack of (unknown) objects, a Franka Panda robot arm, and a suction gripper (Figure \ref{fig:Teaser_YCB_seq_examples}). An image of the scene is fed to SIMstack which outputs meshes and poses. These, along with a target object, are loaded into CoppeliaSim/PyRep \cite{rohmer2013v, james2019pyrep} where 5 virtual cameras are used to create a pointcloud of the visible surface of the stack. The target object's pointcloud is extracted, and grasping locations are sampled based on surface normals. We exhaustively simulate each valid grasp and measure the mean displacement of all objects (excluding the target object) after the grasp has been made. The grasp which produces the lowest mean displacement is chosen to run on the real platform. Note that due to COVID restrictions, we were unable to show the grasps running on the real robot. Two examples of successful grasps in simulations based on SIMstack reconstructions of real object  piles are shown in Figure \ref{fig:robot_demo}, and a demo of the system can be seen in the \href{https://zoelandgraf.github.io/SIMstack/}{supplementary video}.
\vspace{-0.1cm}
\section{Discussion and Conclusion}
We propose a novel method (SIMstack) to generate 3D shape and class-agnostic instance segmentation for multiple stacked objects from a single depth image. We learn a joint shape and instance encoding, trained on a dataset of parametric shapes, randomly assembled under physics simulation. This encoding acts as an intutive physics prior for realistic object stacks, improving reconstruction and segmentation in occluded regions. SIMstack can generate a 3D shape and instance decomposition of a collection of (convex) objects from a single depth view. This output allows for a quick estimate of shape and instance decomposition, which could be used for downstream applications (e.g to initialise a multi-view scanning system to capture more detail~\cite{Wada:etal:CVPR2020}) and allows for fast multi-object reasoning, useful for interactive tasks such as the precise (non-disruptive) grasping we demonstrate. Our method can leverage multi-view information for improved reconstruction which makes it by design a candidate for incremental reconstruction systems.  

We believe our approach can play an important role in rapid scene understanding to help embodied AI systems make physically intuitive interpretations of ambiguous scenes. 
Promising future research directions include integrating RGB information and extending our approach to non-convex objects.
{\small
\bibliographystyle{ieee_fullname}
\bibliography{ms}
}

\clearpage

\begin{appendices}

\paragraph{}
\textit{We provide the following additional content: (A) complementary graphs for our single-view evaluation and additional qualitative results comparing SIMstack to our baselines, (B) a more detailed description of our multi-view optimisation pipeline, (C) a detailed description of our network architectures and (D) some additional qualitative examples showing results on non-convex object scenes and a latent code analysis.}

\section{Single-view evaluation}
\begin{figure}[ht]
\centering
\begin{minipage}[c]{0.49\linewidth}
  \includegraphics[width=1.0\linewidth]{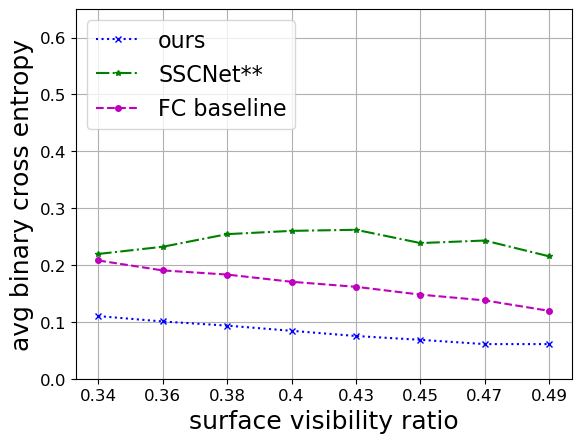}
\end{minipage}
\begin{minipage}[c]{0.49\linewidth}
  \includegraphics[width=\linewidth]{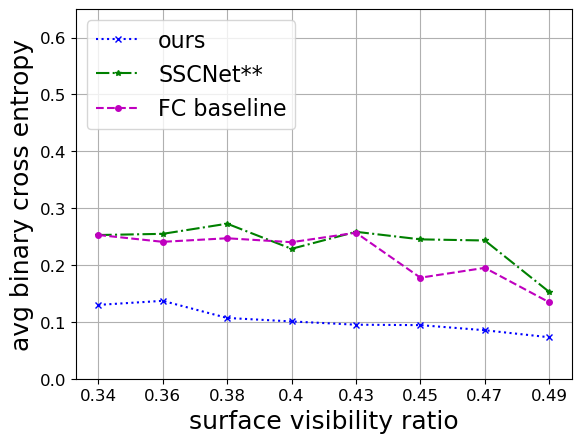}
\end{minipage}
  \caption{\small{We compare our C-VAE against $SSCNet^{**}$ and our fully convolutional (FC) baseline for predicted (expected) voxel occupancy. \textbf{Left}: test dataset of SuperQuadric shapes \textbf{Right}: our YCB object test dataset.}}
\label{fig:expected occupancy}
\vspace{2mm} \hrule \vspace{-0.2cm} 
\end{figure}

\begin{figure}[ht]
\centering
\begin{minipage}[c]{0.49\linewidth}
  \includegraphics[width=1.0\linewidth]{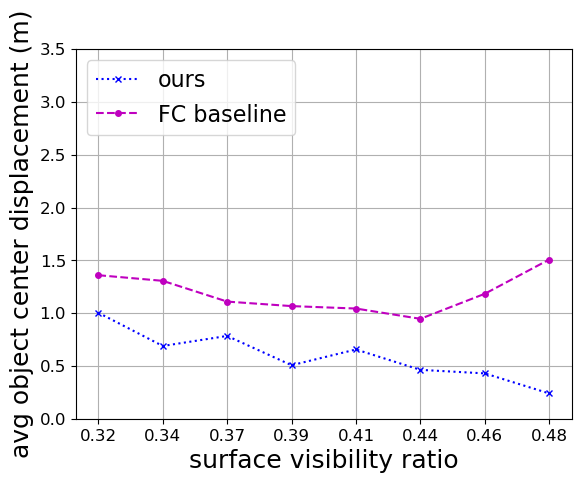}
\end{minipage}
\begin{minipage}[c]{0.49\linewidth}
  \includegraphics[width=\linewidth]{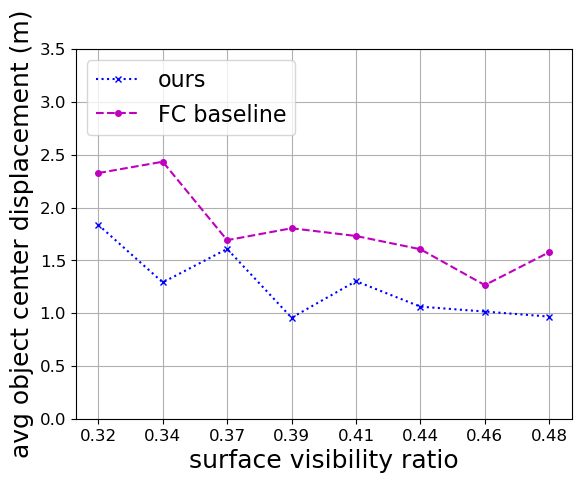}
\end{minipage}
  \caption{\small{We compare SIMstack and our fully convolutional (FC) baseline in terms of stability under physics simulation (10000 steps) by computing the average object displacement per object stack (m). \textbf{Left}: test dataset of SuperQuadric shapes \textbf{Right}: our YCB object test dataset.}}
\label{fig:stability_eval}
\vspace{2mm} \hrule \vspace{-0.2cm} 
\end{figure}

\begin{figure}[ht]
\centering
\begin{minipage}[c]{0.49\linewidth}
  \includegraphics[width=1.0\linewidth]{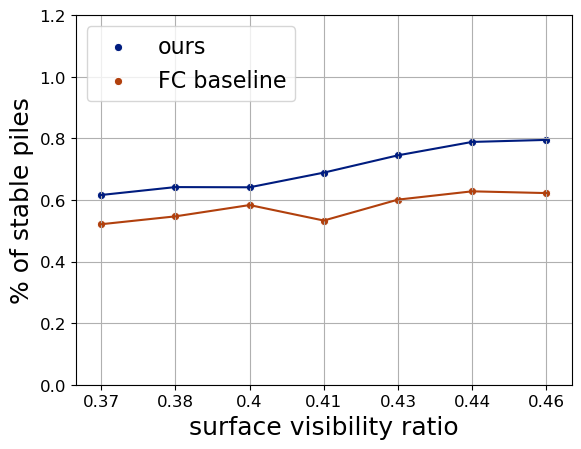}
\end{minipage}
\begin{minipage}[c]{0.49\linewidth}
  \includegraphics[width=\linewidth]{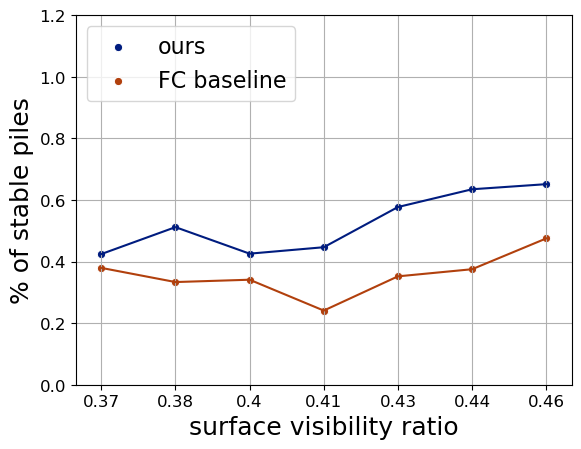}
\end{minipage}
  \caption{\small{Percentage of stable piles generated according to our stability threshold (all objects center displacement stays within $20$cm and all objects orientation change stays within $30^{\circ}$). \textbf{Left}: test dataset of SuperQuadric shapes \textbf{Right}: our YCB object test dataset.}}
\label{fig:stability}
\vspace{2mm} \hrule \vspace{-0.2cm} 
\end{figure}

In addition to our results reported in the main paper, we show the corresponding graphs displaying results per surface visibility ratio for predicted voxel occupancy and stability under physics simulation (Figures \ref{fig:expected occupancy} and \ref{fig:stability_eval}). We also provide an additional evaluation of stability; in the case of bad predictions (unstable decompositions), rolling objects as well as intersecting objects, pushed apart by contact force can cause large object centre displacements. To provide a more intuitive notion of stability, we estimate the percentage of stable piles according to a stability threshold, which we determine through observation. To allow some shuffling of objects, but exclude falling objects, we set the following threshold: we define a stack to be stable if none of its composing objects' centres move by more than $20$cm and all objects' orientation change stays within $30^{\circ}$. We plot the percentage of stable stacks by surface visibility on our test datasets in Figure \ref{fig:stability}.

\begin{figure*}
\includegraphics[width=1.0\linewidth]{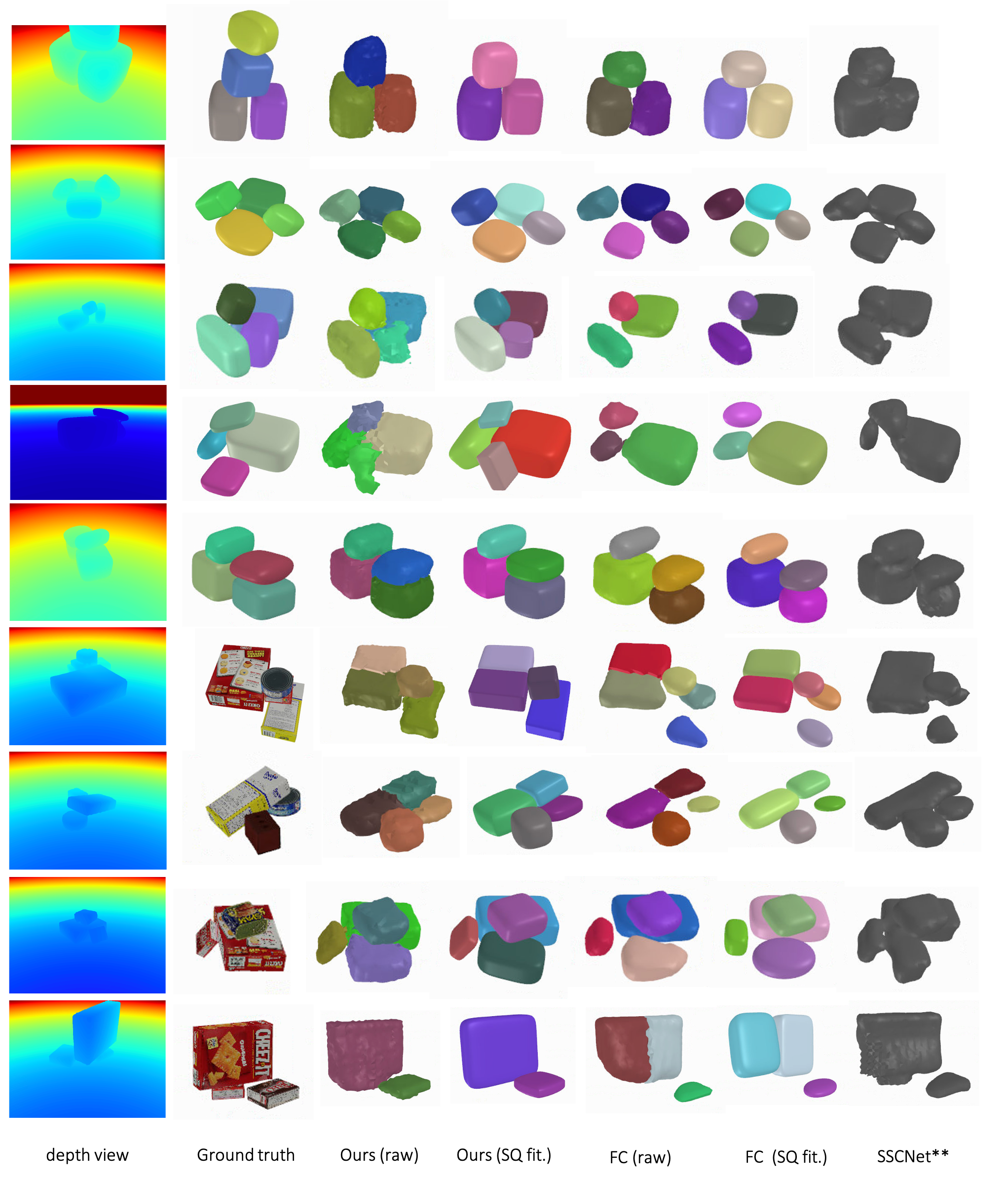}
\caption{Qualitative results comparing SIMstack (Ours) to our fully convolutional (FC) baseline and our SSCNet baseline ($SSCNet^{**}$). We select a random viewpoint for each scene and display a random latent code sample for SIMstack.}
\label{fig:qual_results}
\end{figure*}

We provide additional qualitative results on our Superquadric (SQ) shape test dataset and our YCB object test set, comparing SIMstack to our baselines (see Figure \ref{fig:qual_results}).

\section{Multi-view optimisation}
As shown in the main paper, SIMstack can integrate additional views to improve reconstruction and instance segmentation using \textit{multi-view conditioning} and \textit{multi-view optimisation}. In this section we provide more details on our multi-view optimisation pipeline.
\label{app:differentiable_renderer}
\paragraph{Differentiable Depth Renderer}
\label{sec:Diff_Renderer}
Our differentiable depth renderer is based on SDF ray casting. Similarly to \cite{Jiang_2020_CVPR}, we use sphere tracing 

to render depth. While stepping along each ray, we compute the exact SDF value using trilinear interpolation.

Jiang \etal \cite{Jiang_2020_CVPR} only backpropagate the gradients into the immediate neighbourhood of each ray-surface intersection, which is correct when using a depth-image based loss. Since we use a more precise SDF based loss (Equation \ref{eq:TSDF_Loss}), we need gradients in the entire field of view of the camera. Although there has been recent work on fully differentiable sphere tracing \cite{Liu_2020_CVPR}, we choose a simpler approximation, sampling SDF values at regular intervals along every ray outside the surface to backpropagate gradients within the entire camera frustrum. We estimate our gradients using the binary loss described by Equation (\ref{eq:binary_loss}). Parallelised, our method can render the full cost image at an average runtime of $0.192 s$ at a resolution of $640 \times 480$. In comparison, \cite{Liu_2020_CVPR} render an image of $512 \times 512$ in $0.99s$.

\paragraph{Cost Function}
\label{sec:opt_loss_functions}
Although related approaches use a depth loss to optimise their latents \cite{Jiang_2020_CVPR,Sucar:etal:3DV2020}, we observe that for our TSDF representation using a depth-based loss leads to an incorrect definition at occlusion boundaries.

We design an SDF-based cost function composed of two parts: one describing the loss at the visible surface and one for the visible, unoccupied region of the scene. Let $\pi^{-1}$ be the function which backprojects a pixel $\mathbf{u_{i}}$ of the depth image into the 3D scene and

$I$ is the trilinear interpolation function which obtains the TSDF value at that point. The surface loss is:
\begin{equation}
    L_{\text{surface}} = \sum_{\Omega}{I(\pi^{-1}(\mathbf{u_{i}}))}.
    \label{eq:TSDF_Loss}
\end{equation}
The current TSDF is optimised towards alignment with this surface data, but this loss doesn't constrain on visible regions of empty space. We therefore define an empty space loss which penalises the code if it produces a negative TSDF value in observed empty space. We sample at regular intervals along rays in all regions of observed empty space and define the empty space loss, which has a `space carving' effect: $L_{\text{empty\_space}} = \sum_s L_{\text{empty\_space}, s}$, where:
\begin{eqnarray}
L_{\text{empty\_space}, s} = {\begin{cases}
       {|I(s)|}\quad\text{if } I(s) < 0 \\
       {0} \quad\text{if } I(s) > 0
       
     \end{cases}
     }
     \label{eq:loss}
~.
\label{eq:binary_loss}
\end{eqnarray}
Here $s$ is a sampled TSDF value.
Our final cost function is the simple sum of both losses:
\begin{equation}
    L = L_{\text{surface}} + L_{\text{empty\_space}}.
    \label{eq:LossFunction}
\end{equation}

\paragraph{Optimisation (implementation details)} We use first-order optimisation to optimise our latent code against additional depth images: Once the loss $L$ is computed, we backpropagate the gradients into the voxel-grid using inverse raytracing and trilinear interpolation, paralellised on the GPU. For multiple depth images, we accumulate the gradients of all views. We then leverage PyTorch autograd to backpropagate the gradients through the generative decoder of our C-VAE and use Adam to generate gradient updates.

\paragraph{Runtime} Our multi-view conditioning method's runtime only depends on the time to generate a partial TSDF ($5.2 s$ for 6 views using our TSDF Fusion method) from multiple views as the forward pass time stays constant. It clearly outperforms our multi-view optimisation method which takes $21 s$ and $75 s$ to optimise against $1$ and $6$ views respectively, for 30 iterations.

\section{Network Architecture Details}
\label{app:architecture_details}
\begin{figure}[ht]
\centering
\includegraphics[width=1.0\linewidth]{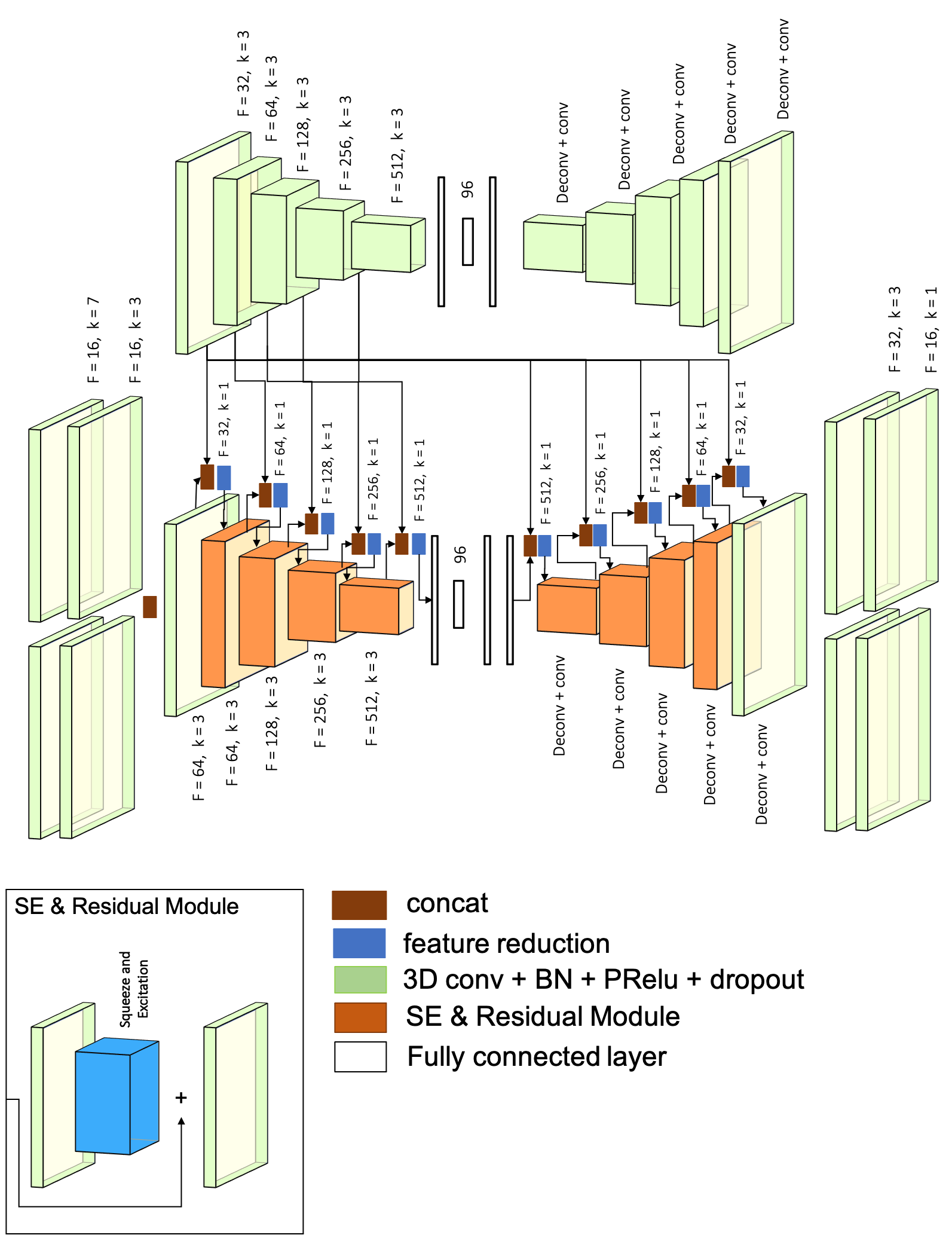}
\caption{C-VAE architecture}
\label{fig:C-VAE}
\end{figure}

\paragraph{C-VAE}
We provide a detailed overview of our C-VAE in Figure \ref{fig:C-VAE}. \textbf{Conditioning AutoEncoder} Our conditioning network is trained as an autoencoder, encoding and decoding the partial TSDF generated from the input depth view. Encoder and Decoder each have 5 convolutional layers with 2 linear layers compressing the feature maps into a 1D bottleneck of $96$. The encoder feature maps are used to condition the encoder and decoder of the shape and instance VAE. \textbf{Shape and Instance VAE} Our VAE's encoder maps input TSDF and instances to a common feature space using two convolutional layers for each modality. The resulting feature maps are concatenated and further compressed using $5$ joint encoding layers. One linear layer maps our 3D feature space to our 1D latent code of size $96$ while $2$ linear layers map it back to 3D. Our VAE 3D decoder mirrors our encoder. We use Batch Normalisation (BN) and PRelu activations in all of our hidden layers.

\begin{figure}[ht]
\centering
\includegraphics[width=1.0\linewidth]{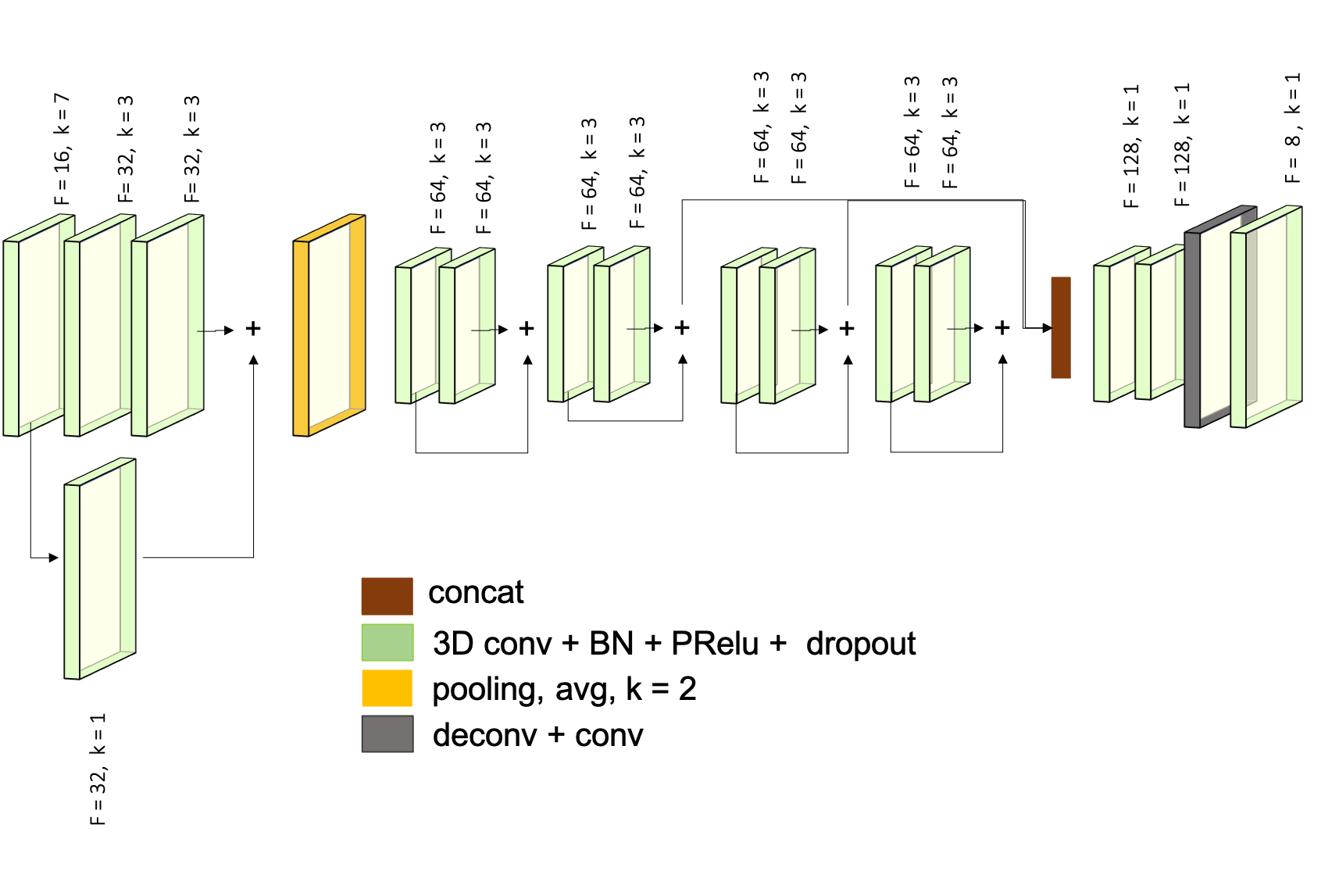}
\caption{$SSCNet^{**}$ architecture}
\label{fig:SSCNet}
\end{figure}

\paragraph{SSCNet Baseline}
We provide a detailed overview of our adapted version of SSCNet in Figure \ref{fig:SSCNet}. To fit this baseline to our task which requires same-resolution output and TSDF prediction, we adapt the SSCNet architecture by (1) remove downsampling in most layers by adding padding (2) adding an upsampling (deconv) layer at the end to generate a prediction at the same resolution as the input and (3) predicting TSDF values instead of semantic labels. 

\begin{figure}[ht]
\centering
\includegraphics[width=0.9\linewidth]{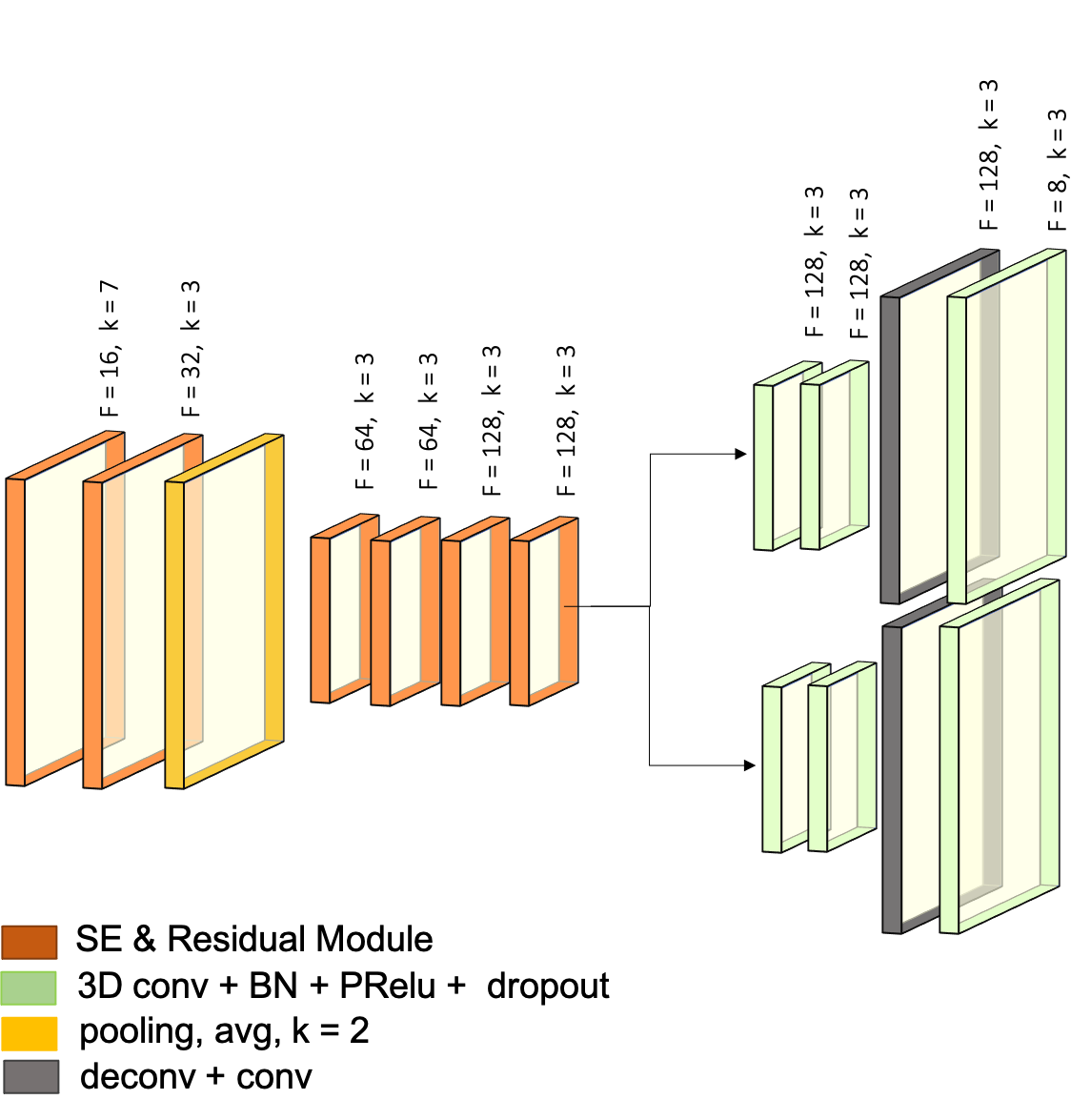}
\caption{FC baseline architecture}
\label{fig:FC_baseline}
\end{figure}

\paragraph{FC Baseline}
Figure \ref{fig:FC_baseline} shows our fully convolutional (FC) baseline architecture. We use a partial TSDF encoder branch with the same SE and Residual units used in our C-VAE, which splits into two task specific decoder branches.

\section{Additional Experiments}
\label{app:additional_experiments}
\subsection{Decomposing more complex scenes}
\begin{figure}[ht]
\centering{
\begin{minipage}[c]{0.99\linewidth}
\includegraphics[width=\linewidth]{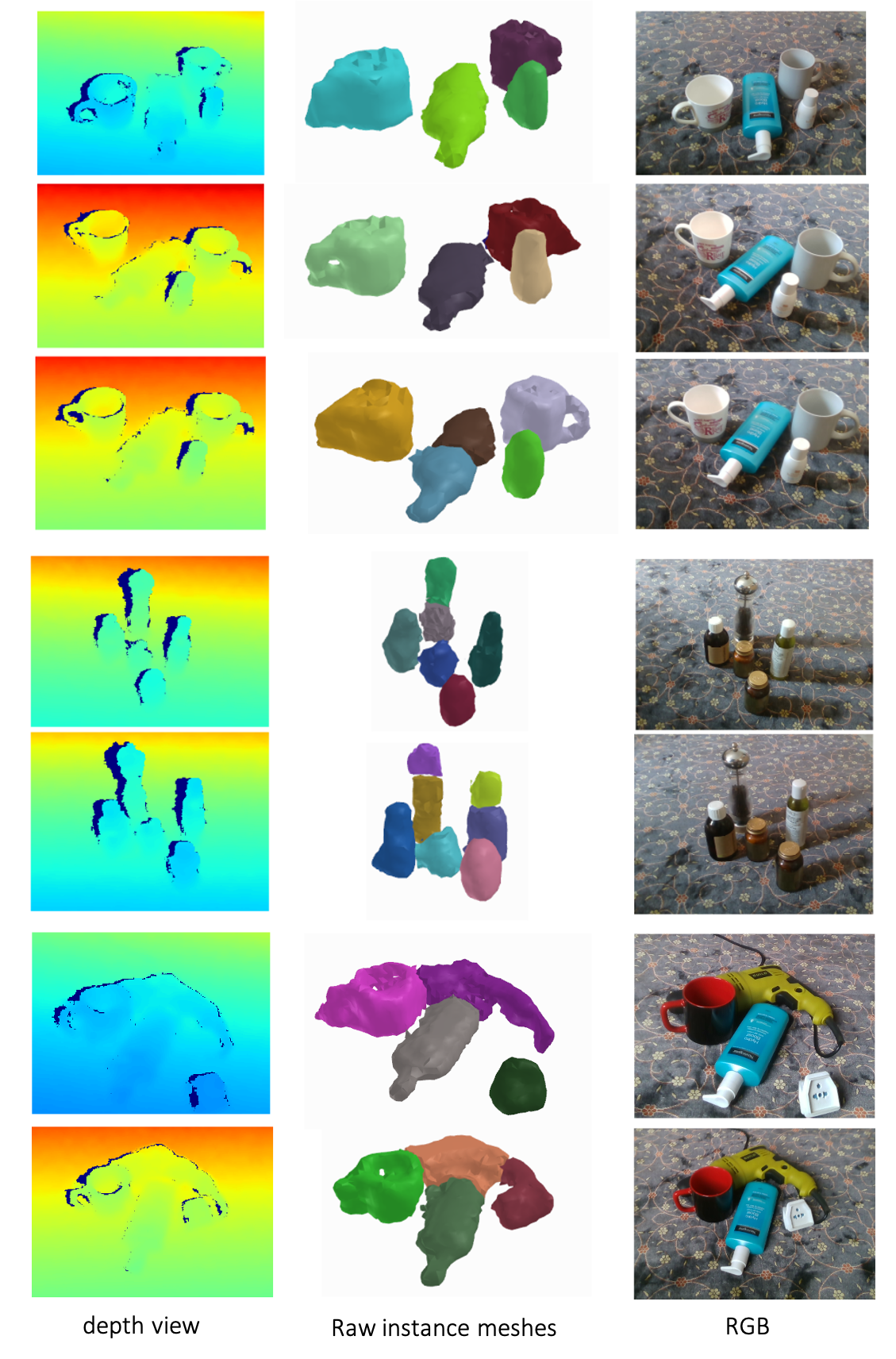}
\end{minipage}
}
  \caption{\small{Qualitative results on real scenes with non-convex objects. The raw mesh segmentation of SIMstack is able to estimate shape and decomposition of scenes with non-convex objects.}}
\label{fig:concave_objects}
\vspace{2mm} \hrule \vspace{-0.3cm} 
\end{figure}

We test our method on real scenes with a collection of non-convex objects and find that although trained on convex shapes only, our approach (excluding parametric fitting) shows some ability to generalise to such scenes. Our reconstructions from different viewpoints for each scene show that although instance decomposition varies across viewpoints and in some cases instances are oversegmented, our C-VAE is able to distinguish individual instances and generate a rough reconstruction of cups, bottles and even part of a drill (see Figure \ref{fig:concave_objects}). This suggests that our method could be extended to more complex scenes with non-convex objects; steps to achieve this could include augmenting the dataset with non-convex shapes and using a different shape refinement step. 

\subsection{Latent Code Analysis}
\begin{figure}[ht]
\centering{
\begin{minipage}[c]{1.0\linewidth}
  \includegraphics[width=\linewidth]{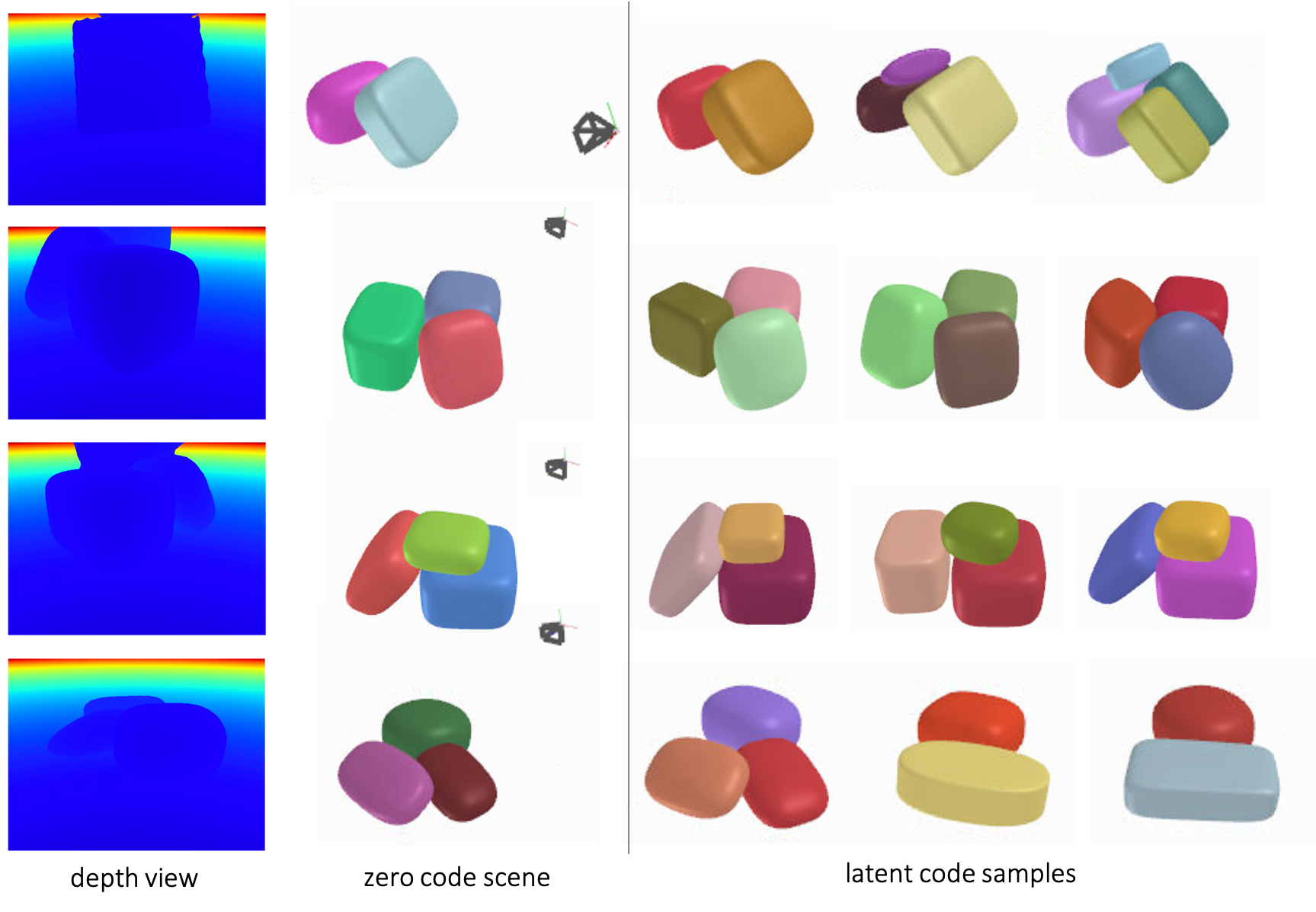}
\end{minipage}
}
\caption{\small{Sampling from the latent code of our C-VAE. Given a single depth image of our SQ test dataset (left), we generate the zero code scene and three latent code samples from our C-VAE.}}
\label{fig:latent_code_samples}
\vspace{2mm} \hrule \vspace{-0.4cm} \end{figure}

In this final section, we provide additional experiments which demonstrate the smoothness and consistency of our joint shape and instance encoding. 
\paragraph{Sampling from the latent space}
Given the design of our C-VAE, sampling from its latent space generates different proposals for occluded regions, while reconstruction of the visible surface stays constant. We show example of this on our Superquadrics test dataset in Figure  \ref{fig:latent_code_samples}. We show the mean scene (zero code scene) along with random latent code samples for a random viewpoint. 

\begin{figure}[ht]
\centering{
\begin{minipage}[c]{0.35\linewidth}
  \includegraphics[width=0.99\linewidth]{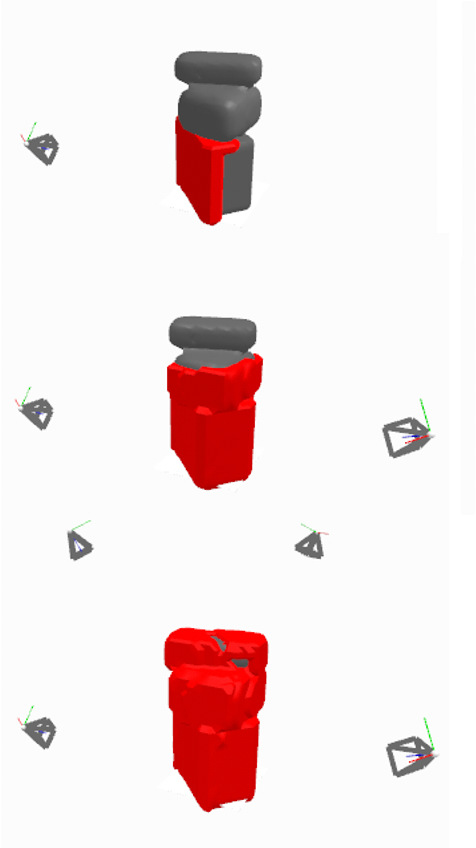}
\end{minipage}
}
\centering{
\begin{minipage}[c]{0.50\linewidth}
  \includegraphics[width=0.95\linewidth]{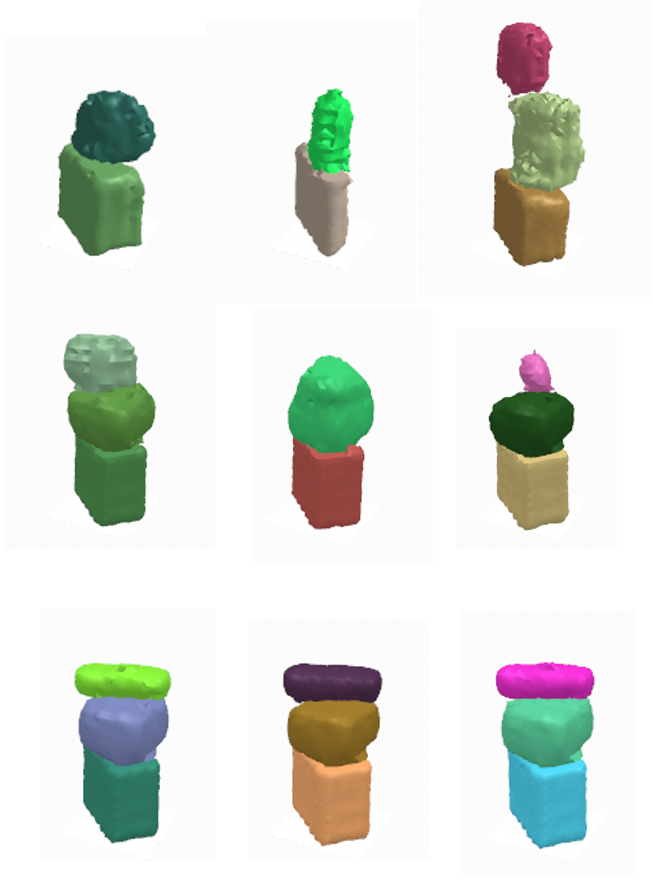}
\end{minipage}
}
 \caption{\small{How sampling variety changes as views are added. \textbf{Left}: Ground truth and visible mesh area overlay (red). \textbf{Right}: 3 random latent space samples (without SQ fitting) for each view; as data is added the samples are increasingly constrained.}}
\label{fig:Multi_view_SAMPLING}
\vspace{2mm} \hrule \vspace{-0.4cm} \end{figure}

\paragraph{Sampling with multi-view information} 
We condition our VAE on depth information to improve reconstruction in visible regions and to allow the latents to focus on occluded regions; sampling from those latents generates a variety of propositions for shape and instance segmentation of those hidden regions. Adding additional views increases the information about 3D space and should show a decreasing variety in latent space samples. We demonstrate this on an example from our Superquadrics test dataset in Figure \ref{fig:Multi_view_SAMPLING}.

\begin{figure}[ht!]
\centering{
\begin{minipage}[c]{1.0\linewidth}
  \includegraphics[width=\linewidth]{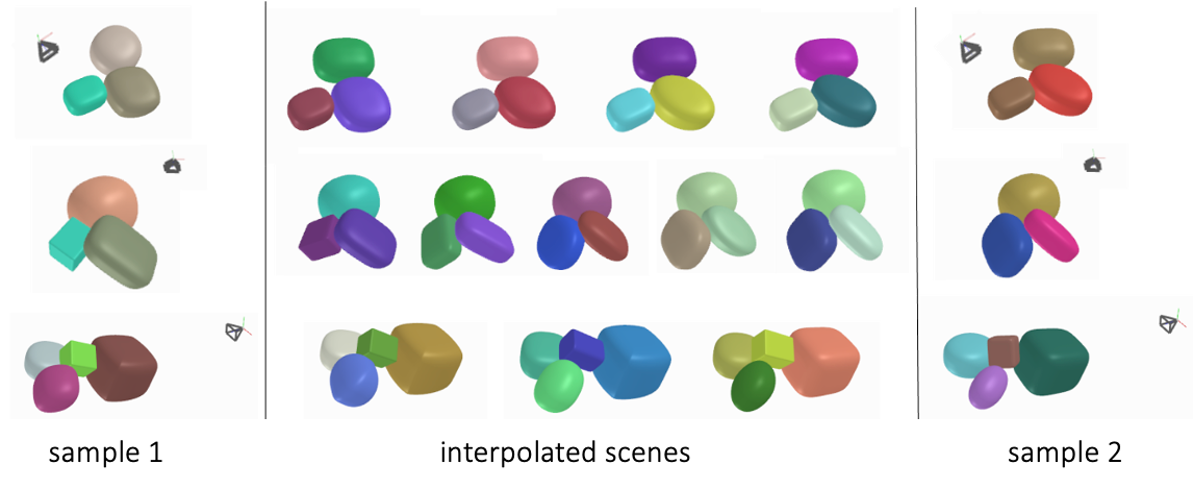}
\end{minipage}
}
\caption{\small{Latent code interpolations between two random latent code samples, conditioned on one view. We show three examples of interpolating between scenes with the \textit{same number of instances} .}}
\label{fig:latent_code_interp_same}
\vspace{2mm} \hrule \vspace{-0.4cm} \end{figure}

\begin{figure}[ht!]
\centering{
\begin{minipage}[c]{1.0\linewidth}
  \includegraphics[width=\linewidth]{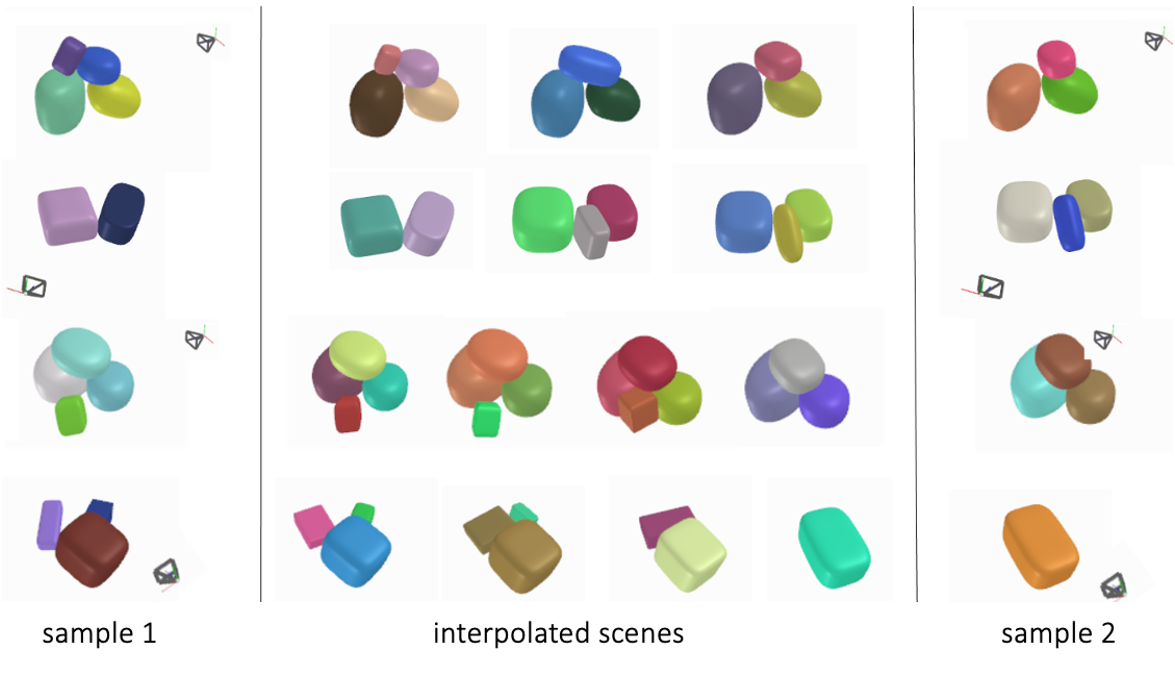}
\end{minipage}
}
\caption{\small{Latent code interpolations between two random latent code samples, conditioned on one view. We show four examples of interpolating between scenes of \textit{varying numbers of instances}  .}}
\label{fig:latent_code_interp_diff}
\vspace{2mm} \hrule \vspace{-0.4cm} \end{figure}

\paragraph{Latent code interpolation}
We qualitatively evaluate the smoothness of our latent code by interpolating between random latent code samples of a depth-conditioned 3D reconstruction. Given a test scene and one viewpoint, we interpolate between two latent code samples to generate intermediary scenes. Our interpolations show a visibly smooth transition between scenes with the same number of instances (see Figure \ref{fig:latent_code_interp_same}) as well as realistic intermediary scenes for interpolations between scenes of varying numbers of instances (see Figure \ref{fig:latent_code_interp_diff}): in the first example, the small object on the top left present in sample 1 becomes smaller, then merges into a long object which then shortens as interpolation approaches sample 2.
\end{appendices}
\end{document}